\documentclass[acmtog,noacm]{acmart}

\renewcommand\footnotetextcopyrightpermission[1]{} 

\usepackage{multirow}
\usepackage{booktabs}
\usepackage{colortbl}
\usepackage[capitalise]{cleveref}

\definecolor{RDcolor}{rgb}{0.5, 0.1, 0.8}

\definecolor{Aycecolor}{rgb}{0.651, 0.165, 0.545}

\definecolor{RevColor}{rgb}{0.0, 0.15, 0.75}

\definecolor{pastelyellow}{RGB}{255, 247, 204}
\definecolor{pastelgreen}{RGB}{220, 255, 220}
\definecolor{pastelblue}{RGB}{220, 235, 255}
\definecolor{pastelpink}{RGB}{255, 220, 235}

\definecolor{pastelyellowdark}{RGB}{255, 235, 170}
\definecolor{pastelgreendark}{RGB}{190, 235, 190}
\definecolor{pastelbluedark}{RGB}{185, 210, 255}
\definecolor{pastelpinkdark}{RGB}{240, 190, 215}

\newcommand{\best}{\cellcolor{pastelgreendark}}
\newcommand{\second}{\cellcolor{pastelgreen}}

\begin{document}

\begin{teaserfigure}
\centering
  \includegraphics[width=0.92\linewidth]{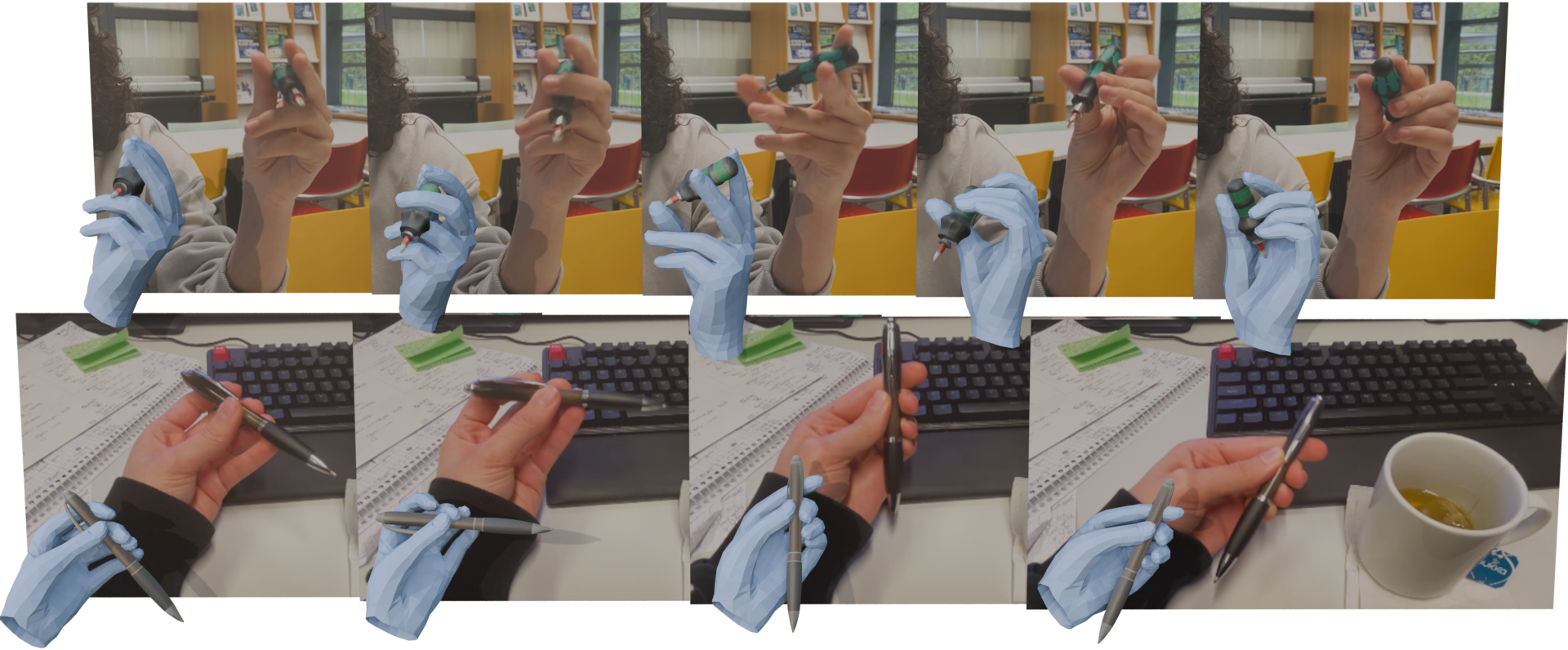}
  \caption{
  \textbf{Grasp in Gaussians (GraG):} 
  Given a single monocular video of a hand interacting with an object, \textit{GraG} reconstructs 3D geometry and pose of the hand and the object. 
  Our method is designed to be efficient for long sequences, and can reconstruct in-the-wild captured examples.
}
  \label{fig:teaser}
\end{teaserfigure}

\author{Ayce Idil Aytekin}
\affiliation{%
  \institution{MPI for Informatics, SIC \& VIA Research Center}
  \country{Germany}
}
\email{aaytekin@mpi-inf.mpg.de}

\author{Xu Chen}
\affiliation{%
  \institution{Google}
  \country{Switzerland}
}
\email{xche@google.com}

\author{Zhengyang Shen}
\affiliation{%
  \institution{Google}
  \country{USA}
}
\email{zyshen@google.com}

\author{Thabo Beeler}
\affiliation{%
  \institution{Google}
  \country{Switzerland}
}
\email{tbeeler@google.com}

\author{Helge Rhodin}
\affiliation{%
\institution{MPI for Informatics}
\country{Germany}
}
\affiliation{%
  \institution{Bielefeld University}
  \country{Germany}
}
\email{helge.rhodin@uni-bielefeld.de}

\author{Rishabh Dabral}
\affiliation{%
\institution{MPI for Informatics, SIC \& VIA Research Center}
  \country{Germany}
}
\email{rdabral@mpi-inf.mpg.de}

\author{Christian Theobalt}
\affiliation{%
\institution{MPI for Informatics, SIC \& VIA Research Center}
  \country{Germany}
}
\email{theobalt@mpi-inf.mpg.de}

\title{Grasp in Gaussians: Fast Monocular Reconstruction of Dynamic Hand–Object Interactions}

\ccsdesc[500]{Computing methodologies~Reconstruction}

\keywords{3D reconstruction, hand-object interaction}

\begin{abstract}
We present \textbf{Grasp in Gaussians (GraG)}, a fast and robust method for reconstructing dynamic 3D hand–object interactions from a single monocular video.
Unlike recent approaches that optimize heavy neural representations, our method focuses on tracking the hand and the object efficiently, once initialized from pretrained large models.
Our key insight is that, given strong pretrained object and hand priors, accurate and temporally stable hand–object motion can be recovered using a compact Sum-of-Gaussians (SoG) representation, revived from classical tracking literature and integrated with generative Gaussian-based initializations.
We initialize object pose and geometry using a video-adapted SAM3D pipeline, then convert the resulting dense Gaussian representation into a lightweight SoG via subsampling. 
This compact representation enables efficient and fast tracking while preserving geometric fidelity, with the pretrained priors and simple geometric/contact losses providing accurate hand-object placement.
For the hand, we adopt a complementary strategy: starting from off-the-shelf monocular hand pose initialization, we refine hand motion using simple yet effective 2D joint, silhouette, depth, and contact losses, avoiding per-frame refinement of a detailed 3D hand appearance model while maintaining stable articulation.
Extensive experiments on public benchmarks demonstrate that GraG reconstructs temporally coherent hand-object interactions on long sequences $4.4\times$--$38.9\times$ faster than prior work while preserving temporally coherent hand-object motion.
Project page:
\href{https://aidilayce.github.io/GraG-page/}{GraG-page}.
\end{abstract}

\maketitle

\section{Introduction}
\label{sec:intro}
Our hands enable us to perform countless interactions with the physical world, from grasping everyday objects to manipulating tools with precision.
Reconstructing such dynamic 3D interactions is a fundamental modeling task with crucial applications in robotics, where learning from human demonstration is key, and in augmented reality, for seamless virtual-physical interaction.
Our goal is to reconstruct hand-object interaction (HOI) from a monocular video, which involves estimating the 3D geometry and pose of the hand and the object.
In this work, we focus on the foundational setting of manipulating a single object with the dominant hand, which covers a vast majority of functional daily tasks.
Yet, performing such reconstruction is challenging, as heavy occlusions and depth ambiguity make it a fundamentally ill-posed problem.
The challenge is further exacerbated when runtime considerations are introduced.
\par
In the absence of clear, unoccluded visual cues, it becomes important to introduce appropriate priors to curb the ill-posedness of the task.
However, past methods have done so with a variety of strong assumptions that often limit their practical applicability to open-world and unconstrained scenarios with complex backgrounds.
For instance, many methods require a pre-scanned 3D object template~\cite{fan2023arctic, hasson2020leveraging, hasson2021towards, yang2021cpf}, making them unable to reconstruct novel unseen objects.
While some template-free methods exist, they are often trained on datasets with a limited number of object instances~\cite{hasson2019learning, karunratanakul2020grasping, ye2022s} or are restricted to specific, predefined object categories~\cite{ye2023diffusion}, leading to poor generalization and inaccurate geometry for out-of-distribution objects. 
Another line of work, in-hand object scanning~\cite{hampali2023hand, huang2022reconstructing, zhong2024color}, can reconstruct novel objects but typically assumes a rigid hand pose, failing to capture dynamic, dexterous articulations.
Recent video-based approaches such as HOLD~\cite{fan2024hold} and BIGS~\cite{on2025bigs} enable category generalization by optimizing implicit surfaces or Gaussian representations.
However, optimizing such volumetric representation across several frames is painstakingly slow ($\sim$10 hours for HOLD, $\sim$4 hours for BIGS for a video with 100 frames), thereby limiting their use on long, in-the-wild video reconstruction.
\par
To overcome these restrictions, we examine what modern priors can offer and what fast tracking requires.
Recent image-to-3D models~\cite{xiang2025trellis, lai2025hunyuan3d} can produce plausible object geometry from a single view, but na\"ively re-running such models frame-by-frame yields jittery poses and inaccurate shape under occlusion.
In comparison, classical model-based human/object tracking formulations are fast and stable, but have historically required carefully engineered templates or multi-view capture~\cite{zhong2000object, sridhar2014real}, which cannot directly generalize to hand-object reconstruction.
In this paper, we show that we can combine the strengths of both worlds by \textit{using a generative 3D prior to obtain a canonical object, and then performing lightweight tracking in a compact representation.}
\par
Our method, dubbed \textit{Grasp in Gaussians (GraG}), operates in three stages.
First, we select a small set of informative and diverse keyframes from the input video and reconstruct a canonical object using MV-SAM3D~\cite{mv-sam3d}, a multi-view extension of SAM3D~\cite{sam3d}.
MV-SAM3D represents the object as a set of canonical shape tokens (shared across views), which we decode once to obtain a dense 3D Gaussian representation of the object.
Second, we extend SAM3D to videos by freezing the canonical shape and estimating per-frame object pose (rotation, translation, and scale) with temporal guidance in its latent space, which suppresses jitter and prevents occlusion-induced drift.
While dense 3D Gaussians are expressive, directly optimizing a high-resolution set of Gaussians across all frames is unnecessary for tracking and becomes computationally expensive.
At the same time, the appearance generated by image-to-3D models like SAM3D is often too inaccurate to be used for photogrammetric tracking objectives.
Therefore, we sparsify the decoded dense Gaussians into a compact Sum-of-Gaussians (SoG) model~\cite{sog} and optimize a differentiable alignment objective that matches its 2D projection to an image SoG constructed from each frame via quad-tree color clustering.
For the hand, we refine a Dyn-HaMR~\cite{yu2025dyn} initialization using 2D joint, depth, silhouette, and contact priors, achieving stable articulation without heavy optimization.
\par
GraG adopts the reconstruction paradigm of having a strong initial prior followed by efficient tracking without additional training.
Such a design makes our method particularly well-suited for long, unconstrained videos.
Experimental results confirm that GraG achieves state-of-the-art performance on standard benchmarks and generalizes robustly to in-the-wild videos with unseen objects and diverse scenes (\textit{c.f.}~\cref{fig:teaser}).
We further demonstrate the practical applicability of GraG in robotic settings. Specifically, we use GraG to reconstruct hand–object interactions from real human demonstrations and use them as goal poses within a tracking-based control policy.
Our method is highly efficient as we reduce the runtime of prior works from $\sim3$-$10$ hours to $\sim$16 minutes on long sequences. 
Our contributions are threefold:
\begin{itemize}
    \item We propose a fast monocular HOI reconstruction system that leverages pretrained object and hand priors to scale to long, unconstrained videos while drastically reducing runtime.
    \item We extend SAM3D to videos by freezing the canonical shape and tracking only per-frame pose with temporal guidance, improving stability under occlusion.
    \item We introduce compact occlusion-aware SoG tracking for Gaussian-based object assets by sparsifying dense Gaussians and aligning projected object SoG to an image SoG built via quad-tree clustering.
\end{itemize}

\section{Related Work}
\label{sec:related_work}

\paragraph{3D asset generation}
Reconstructing 3D objects from images is a core problem in computer vision.
While traditional multi-view methods like Structure from Motion (SfM) \cite{tomasi1992shape, scharstein2002cw, hartley2011mvgeo, szeliski2010sw} are effective, their performance degrades significantly in videos with heavy occlusion and limited viewpoints.
To address these limitations, recent work has shifted towards learning-based, single-image 3D generation, trained on large-scale datasets like Objaverse~\cite{deitke2023objaverse}.
Diffusion models, in particular, have emerged as the state-of-the-art for generating high-fidelity 3D assets from a single 2D image~\cite{xiang2025trellis, sam3d}.
Recently, SAM3D~\cite{sam3d} proposes pose estimation together with 3D asset generation by separating pose and shape space of the object in the image, resulting in high-quality posed 3D objects. 
To obtain a strong geometric prior for the object, we employ the multi-view extension of SAM3D.

\paragraph{Gaussian tracking}
Gaussian primitives were used for human~\cite{plankers_articulated_2003,Moeslund2006,sog,Elhayek2015,rhodin2016egocap}, hand~\cite{Bretzner2002,Sridhar2015}, and object tracking~\cite{Ren2014} well before modern Gaussian Splatting (GS)~\cite{kerbl3Dgaussians}.
SoG~\cite{sog} is especially useful because representing both the image and 3D template with Gaussians enables efficient matching via closed-form Gaussian overlap.
Later tracking methods improved occlusion handling with ray tracing~\cite{rhodin2015versatile,rhodin2016egocap}, contour tracking~\cite{rhodin2016general}, alpha blending~\cite{kerbl3Dgaussians}, point visibility~\cite{rajivc2025multi}, or dynamic surface tracking~\cite{zheng2025gaustar}.
However, these methods often rely on multi-view or depth input, or optimize dense photorealistic Gaussian renderings~\cite{cong2025dytact, liugeneralizable}.
Opposed to increasing photorealism with GS, we revisit the SoG representation, as it alleviates sorting Gaussians back-to-front (for alpha blending) and reduces the computation cost with the sparse Gaussian image representation.

\paragraph{Monocular hand tracking under object occlusion}
Hand-object occlusion is also a central challenge in monocular hand mesh recovery.
HandOccNet~\cite{park2022handoccnet} injects hand information into occluded image regions with transformer modules, H2ONet~\cite{xu2023h2onet} fuses non-occluded finger and orientation cues across nearby frames, and HaMeR~\cite{pavlakos2024hamer} uses a large transformer-based model for robust monocular hand reconstruction.
Building on HaMeR, Dyn-HaMR~\cite{yu2025dyn} recovers temporally coherent hand motion and can use external camera intrinsics and extrinsics, making it a suitable initialization for our camera-registered hand trajectory.
These works motivate using a strong hand prior under occlusion, but they recover only the hand and do not estimate the interacting object's geometry, scale, or trajectory.
GraG uses monocular hand estimates as geometric anchors while reconstructing the object and its hand-relative motion over time.

\paragraph{Hand-object reconstruction}
Dynamic HOI reconstruction is difficult due to severe mutual occlusion, fast object motion, and the absence of object templates in the wild.
While datasets such as DexYCB~\cite{chao2021dexycb} provide synchronized multi-view RGB-D captures for controlled evaluation, monocular video reconstruction requires reasoning under depth ambiguity.
Early template-based or depth-dependent methods~\cite{cao2021reconstructing, corona2020ganhand, liu2021semi, tekin2019h+o, yang2021cpf} therefore have limited applicability outside controlled settings.
Recent single-image methods such as EasyHOI~\cite{liu2025easyhoi} and FollowMyHold~\cite{aytekin2025follow} improve generalization by composing foundation models, but do not exploit temporal cues from videos.
Video-based methods such as HOLD~\cite{fan2024hold} and BIGS~\cite{on2025bigs} optimize implicit or Gaussian scene representations, while MagicHOI~\cite{wang2025magichoi} adds generative priors to complete occluded regions.
However, these approaches remain computationally expensive and sensitive to initialization (commonly SfM-style bootstrapping~\cite{sarlin2019coarse, sarlin2020superglue}) under occlusion, textureless objects, and long sequences.
Concurrent work, AGILE~\cite{shi2026agile}, addresses these issues by using agentic generation to obtain complete, high-texture, simulation-ready objects and then tracking them without SfM, while ForeHOI~\cite{chen2026forehoi} reconstructs the object geometry from interaction videos in a feed-forward manner.
In contrast, GraG focuses on efficient long-video HOI reconstruction: we reconstruct a canonical object from selected keyframes and then recover temporally stable hand-object motion with compact occlusion-aware SoG tracking, emphasizing fast hand-relative object pose recovery rather than photorealistic texture completion.

Another line of work adds physics to the reconstruction. 
\citet{hu2022physical} recover the interaction and its contact forces with an explicit physics solver from a single RGB-D sensor, and their follow-up, HOIC~\cite{hu2024hand} uses deep reinforcement learning in a physics simulator with object compensation control and a surface-contact model to further improve physical correctness.
These methods highlight the benefit of explicit physical reasoning, but address RGB-D reconstruction with physical motion/contact modeling. 
We instead keep tracking cheap with lightweight contact and penetration priors for efficient monocular tracking and leave stronger physical constraints to future work.
\begin{figure}[t]
    \centering
    \includegraphics[width=1.0\linewidth]{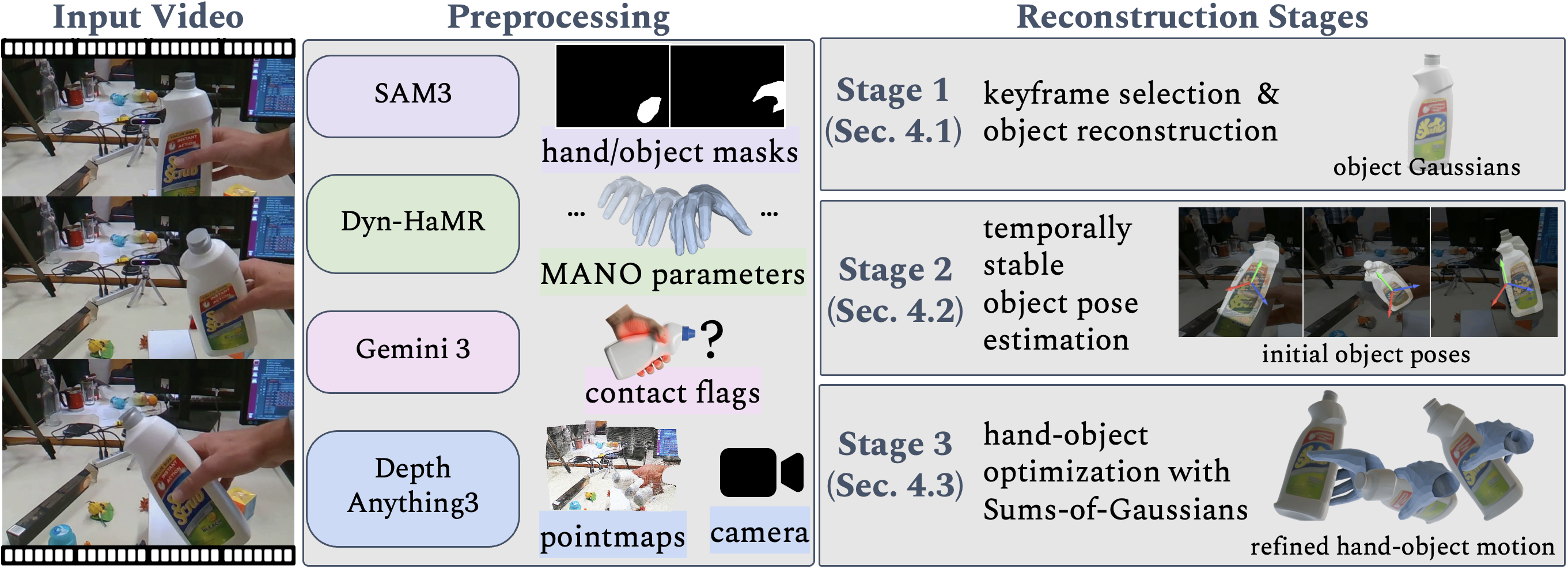}
    \caption{
\textbf{Method Overview.} 
Given a monocular video, we recover per-frame hand-object poses and geometry.
We first preprocess the video to obtain masks, initial hand poses, per-frame binary contact information, pointmaps and camera intrinsics/extrinsics.
Preprocessing is followed by three stages: Stage 1~(\cref{sec:stage1_mvsam3d}) for keyframe selection and object reconstruction, Stage 2~(\cref{sec:stage2_temporal_sam3d}) for estimating a temporally stable initial object trajectory (and scale) by adapting SAM3D to videos, and Stage 3~(\cref{sec:stage3_sog}) for hand and object pose and scale refinement via Sums-of-Gaussians tracking.
}
\label{fig:overview}
\end{figure}

\section{Background: SAM3D}
Our approach to initialize object poses is built upon SAM3D~\cite{sam3d},which is a generative model that reconstructs a 3D object from a single RGB image $I$, object mask $M$, and an optional scene pointmap $P$. Unlike previous methods that predict geometry in an implicit orientation~\cite{xiang2025trellis, lai2025hunyuan3d}, SAM3D explicitly decouples the canonical \textit{shape} $S$ and texture $F$ from the camera-space \textit{layout} $(R,\tau,s)$, which represents the object's orientation, translation, and scale.

Concretely, the model approximates the conditional joint distribution $p(S,F,R,\tau,s \mid I, M, P)$ using a structured multi-modal transformer. Within this architecture, the object is represented by separate canonical shape tokens $Z^S$ and low-dimensional layout tokens $Z^P$. Modality-specific projection layers map these tokens into a high-dimensional feature space, allowing cross-modal attention mechanisms to resolve the final geometry and pose.
\section{Method}

Given a monocular RGB video $\mathbf{I}=\{\mathbf{I}_t\}_{t=1}^{N}$ of a hand interacting with a rigid object, our goal is to recover temporally coherent 3D geometry and motion of the hand and the object.
This task is highly underconstrained due to depth ambiguity and hand-object occlusions.
To tackle it, we leverage the strong data-driven priors provided by generative 3D models like SAM3D and integrate them into the highly efficient classical tracking approach based on Sum-of-Gaussians. 
\par
Fig.~\ref{fig:overview} illustrates the overall schema of our method.
Briefly, we first obtain a reliable canonical object representation from the video (\cref{sec:stage1_mvsam3d}), then estimate per-frame object pose (\cref{sec:stage2_temporal_sam3d}), and finally refine both hand and object trajectories with an explicit, occlusion-aware tracking objective (\cref{sec:stage3_sog}).
This design is motivated by the observation that heavy per-frame 3D reconstruction is unnecessary once a strong canonical prior is available, whereas lightweight tracking can robustly propagate motion through ambiguous and occluded frames.

As preprocessing, we extract hand and object masks $M_t^h$ and $M_t^o$ using SAM3~\cite{sam3}, where object masks are prompted with labels from a large VLM~\cite{v2025glm}. We estimate camera intrinsics/extrinsics $(K_t, T_t)$ and pointmaps $P_t$ with DepthAnything3~\cite{depthanything3}. These camera parameters are then provided as external inputs to Dyn-HaMR~\cite{yu2025dyn} to initialize the hand trajectory. Because Dyn-HaMR regresses the MANO-based hand pose directly within the coordinate frame defined by $(K_t, T_t)$, the hand and object geometries are natively expressed in the same coordinate system, eliminating the need for post-hoc spatial alignment. Finally, we obtain per-frame contact flags $c^{ho}_t \in \{0, 1\}$ using Gemini 3~\cite{gemini}.
Throughout our discussion, we use superscripts $o$ and $h$ to denote object and hand related variables, respectively.

\subsection{Stage 1: Keyframe Selection and Object Reconstruction}
\label{sec:stage1_mvsam3d}
Since our system's input is a video, multiple views of the same object are generally available over time, but occluded by the hand.
Compared to single-frame reconstruction, these views provide complementary visual cues that can significantly improve object shape estimation.
To use this information, and be efficient, we select a compact set of $K$ keyframes $\mathcal{V} = \{\mathcal{V}_k\}_{k=1}^K \subset \mathbf{I}$ that are informative and diverse.
These video keyframes are treated as multi-view inputs and fed to MV-SAM3D~\cite{mv-sam3d}, a multi-view extension of the state-of-the-art image-to-3D model SAM3D~\cite{sam3d}, yielding canonical object shape tokens and dense 3D Gaussians.

\begin{figure}[t]
  \centering
  \includegraphics[width=1.0\linewidth]{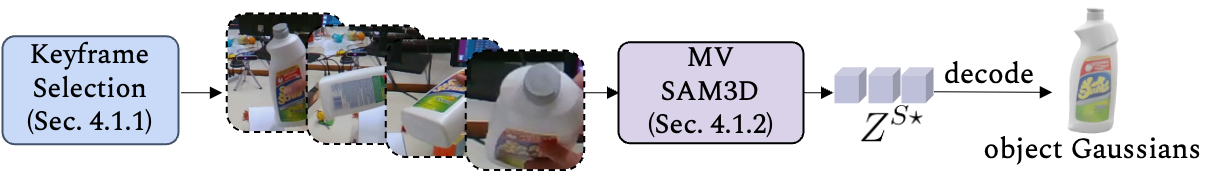}
  \caption{\textbf{Stage 1} reconstructs a canonical object with MV-SAM3D by selecting keyframes and decoding shape tokens into dense Gaussians (\cref{sec:stage1_mvsam3d}).
  }
  \label{fig:stage1}
\end{figure}

\subsubsection{Keyframe selection}
To select informative and diverse views, we adopt the saddle-balanced selection strategy from Depth Anything 3 (DA3)~\cite{depthanything3} using DINOv2 \texttt{[CLS]} features~\cite{dinov2}. 
We compute a balance cost based on normalized feature statistics (mean magnitude, variance, and inter-frame similarity) to filter out unstable or weak activations~\cite{darcet2024vision}. 
Finally, a subset of $K$ keyframes is selected greedily by trading off this stability cost against a diversity penalty to eliminate redundancy (\cref{fig:stage1}). We provide more details in Section 10.4 of the Supplementary Material.

\subsubsection{Object reconstruction}
The K keyframes
selected above are then used as input to MV-SAM3D~\cite{mv-sam3d}.
%
%
The output shape tokens, $Z^{S\star}$, from MV-SAM3D act as a canonical object representation, which we later decode into Gaussians and sparsify for fast tracking~(\cref{sec:gauss_to_sog}).

\subsection{Stage 2: Temporally Stable Object Pose Estimation}
\label{sec:stage2_temporal_sam3d}
\begin{figure}[t]
  \centering
  \includegraphics[width=1.0\linewidth]{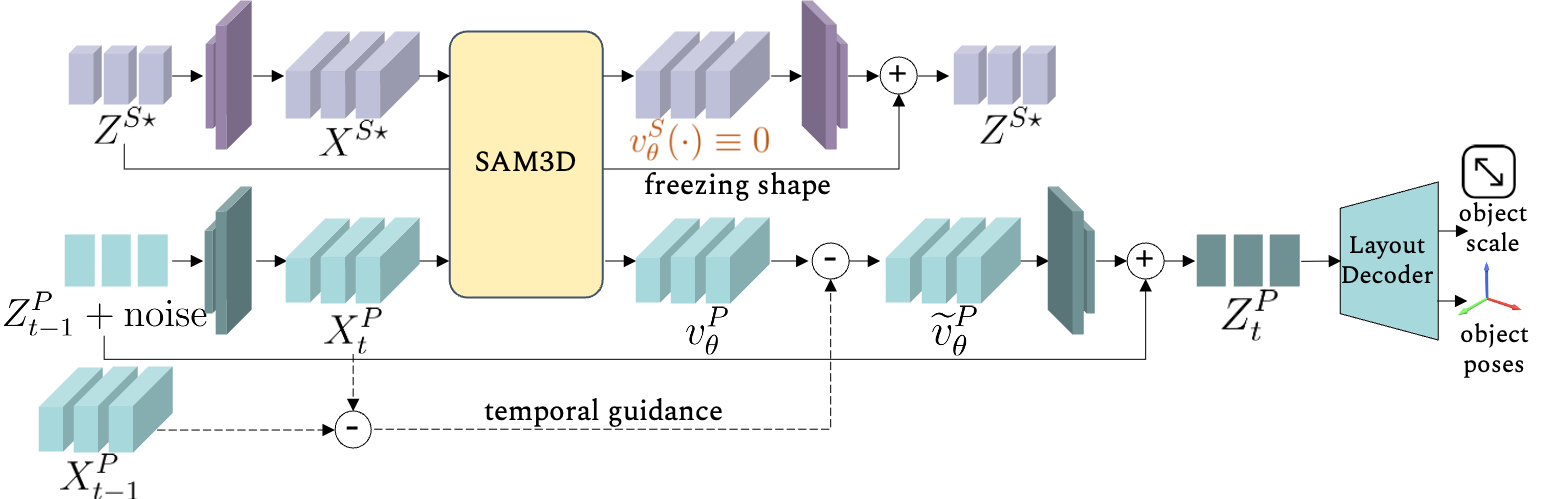}
  \caption{\textbf{Stage 2: Temporally stable SAM3D pose estimation.} Data flow for per-frame pose estimation. \textbf{Top (Shape):} Canonical shape tokens $Z^{S\star}$ are projected to latents $X^{S\star}$ and processed by the transformer. The predicted shape velocity is zeroed to keep the geometry unchanged. \textbf{Bottom (Pose):} Previous frame's layout tokens $Z^P_{t-1}$ are projected to latents $X^P_t$. The predicted velocity $v^P_\theta$ is modified by a temporal guidance term (Eq.~\ref{eq:temporal_guidance}) derived from the difference between current and previous frame latents ($X_{t-1}^P$). The guided velocity $\tilde{v}^P_\theta$ is mapped back to token space and added to the previous frame's tokens to produce $Z^P_t$ for decoding.}
  \label{fig:stage2}
\end{figure}

In Stage~2 (\cref{fig:stage2}), we estimate the per-frame object layout $(R^o_t,\tau^o_t,s^o)$ while forcing the shape to remain constant over time. Following SAM3D's conditional flow matching~\cite{sam3d}, we initialize each frame with shared shape tokens $Z^{S}_t \leftarrow Z^{S\star}$ and zero-out the shape velocity field ($v^{S}_\theta(\cdot) \equiv 0$). This freezes the canonical shape while allowing the layout to evolve via cross-modal attention. This constraint is crucial for hand-object interactions, preventing severe occlusions and rapid appearance changes from corrupting the reconstruction.
We illustrate this stage in \cref{fig:stage2}.

\subsubsection{Temporal guidance on pose latents}
Even with a frozen shape, per-frame layout estimation can jitter or even flip orientation under weak visual cues such as symmetric objects and heavy occlusion.
We therefore bias the per-frame layout velocity toward temporal consistency using a temporal guidance term.
Let $X^{P}_t$ be the current pose/layout latents projected using an MLP from $Z^{P}_t$ and $X^{P}_{t-1}$ the previous frame's latents.
We modify the predicted velocity by
\begin{equation}
\tilde{v}^{P}_\theta \;=\; v^{P}_\theta \;-\; \lambda_{\text{temp}}\,(X^{P}_t - X^{P}_{t-1}),
\label{eq:temporal_guidance}
\end{equation}
applied to translation and rotation latents.
Intuitively, Equation~\eqref{eq:temporal_guidance} discourages abrupt latent changes unless strongly supported by the current observation.
At the end of Stage~2, we obtain an initial object trajectory $\{R^o_t,t^o_t,s^o\}_{t=1}^{T}$.

\subsection{Stage 3: Hand--Object Tracking with SoG}
\label{sec:stage3_sog}
\begin{figure}[t]
  \centering
  \includegraphics[trim=5 10 5 5, width=1.0\linewidth]{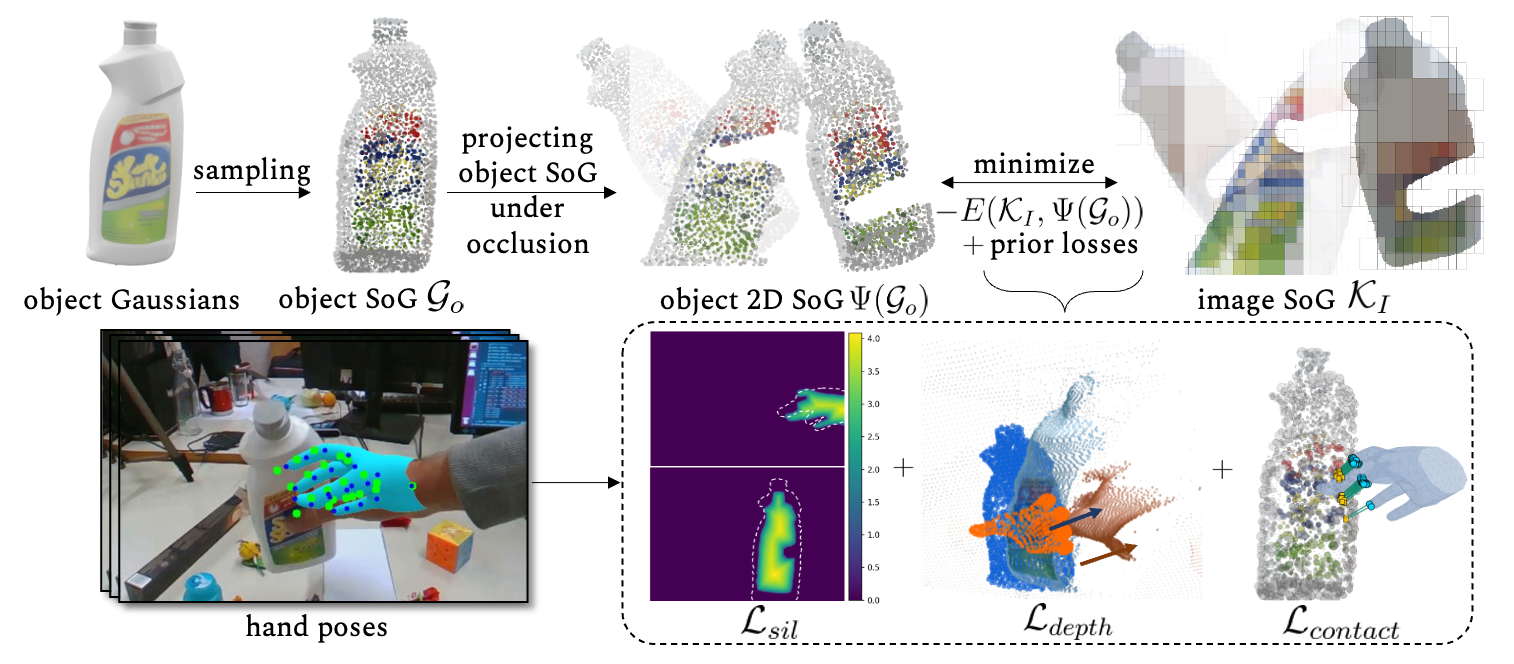}
  \caption{\textbf{Stage 3} refines hand and object motion using an explicit tracking objective: we sparsify the dense Gaussians into a compact SoG and perform hand-aware SoG alignment with lightweight priors including silhouette loss $\mathcal{L}_{sil}$ to represent the distance to the silhouette edges, depth loss $\mathcal{L}_{depth}$ between estimated pointmap medians and hand-object medians, and contact loss $\mathcal{L}_{contact}$ between hand fingertips and object surface (\cref{sec:stage3_sog}). For the silhouette loss, we visualize the $log(DT_{hand}+1)$ and $log(DT_{object}+1)$.
  }
  \label{fig:stage3}
\end{figure}

So far, we have reconstructed the object shape with MV-SAM3D and arrived at an initial estimate of the object trajectory using temporal guidance on pose latents.
Our formulations have been agnostic to the presence of the hand mesh.
Additionally, the estimated object geometry is not guaranteed to be positioned appropriately in the scene.
Therefore, we refine the object trajectory in short windows $\mathcal{W}$ by solving a compact tracking objective.
At a high level, we maximize an image alignment energy $E$ between a set of object Gaussians and the current frame approximated by Gaussians, while regularizing with lightweight geometric and interaction priors:
\begin{equation}
\begin{split}
\min_{\mathcal{X}}
\;&\sum_{t\in\mathcal{W}} -E_t
+
\lambda_{\text{sil}}\mathcal{L}_{\text{sil}}
+
\lambda_{\text{depth}}\mathcal{L}_{\text{depth}} \\
&\quad+
\lambda_{\text{j2d}}\mathcal{L}_{\text{j2d}}
+
\lambda_{\text{contact}}\mathcal{L}_{\text{contact}}
+
\lambda_{\text{smooth}}\mathcal{L}_{\text{smooth}}.
\label{eq:final_obj}
\end{split}
\end{equation}
 where $\mathcal{L}_{\text{sil}}$ aligns the projected object and hand silhouettes with their image masks, $\mathcal{L}_{\text{depth}}$ anchors metric depth and couples object depth to hand depth in contact frames, $\mathcal{L}_{\text{j2d}}$ is the hand-joint reprojection loss, $\mathcal{L}_{\text{contact}}$ encourages physically plausible hand-object contact, and $\mathcal{L}_{\text{smooth}}$ suppresses temporal jitter.
$\mathcal{X}=\{R^o_t,\tau^o_t,s^o,\boldsymbol{\Theta}_t,\allowbreak R^h_t,\tau^h_t,s^h\}_{t\in\mathcal{W}}$ denotes the variables optimized in window $\mathcal{W}$.
We illustrate our optimization approach in \cref{fig:stage3}.
It is worth noting that Stage 3 refines only the poses, not geometry; the canonical object shape from Stage 1 and the hand shape from Dyn-HaMR stay fixed.
We optimize only the per-frame poses and translations, together with one global object scale and one global hand scale per sequence.

Critical to our approach is the image-object alignment term $E_t$, which is computed using Sum-of-Gaussians (SoG) tracking.
By representing the object with a compact set of 3D Gaussians and encoding the masked object image with a compact set of 2D Gaussians, SoG tracking seeks to maximize their continuous overlap similarity.
The SoG energy provides the main visible-object area matching signal; the silhouette term is only a guard against degenerate support, not a separate replacement for SoG alignment.
We next define the main ingredients of $E_t$.
For each frame $t$, we (i) build an \emph{image SoG} from the image inside the object mask, and (ii) project a compact set of object Gaussians (\emph{object SoG}) into the image under the current pose.

\subsubsection{Image SoG}
\label{sec:image_sog}
\begin{figure}[t]
  \centering
   \includegraphics[width=1.0\linewidth]{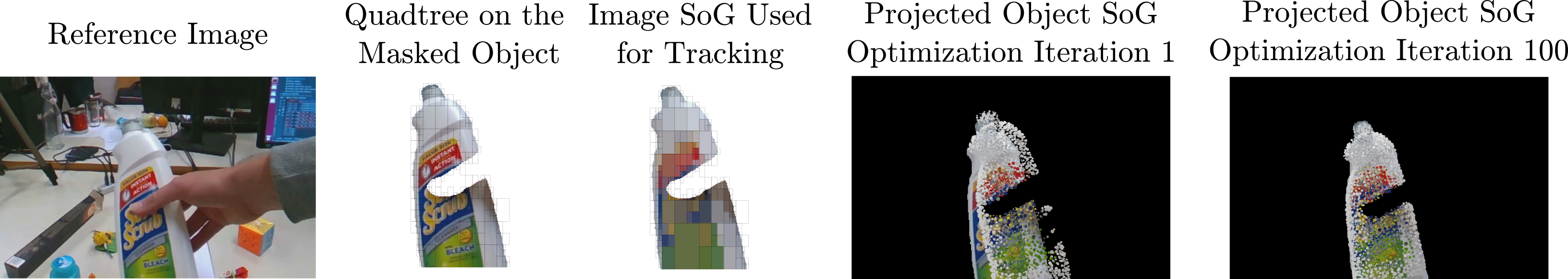}
   \caption{\textbf{Approximating the image as Gaussians and projected object SoG.} From left to right: Reference frame, quad-tree build over the object area, image SoG used for tracking  where each square represents a Gaussian with the same color, projected object SoG before optimization, projected object SoG at the end of optimization.
   }
   \label{fig:image_sog}
\end{figure}

Given an image $I_t$, we seek a compact SoG $\mathcal{K}_{I_t}$ that represents coherent pixel regions inside the object mask $M^o_t$.
A naive construction would assign one Gaussian to each pixel, but this is prohibitively expensive.
Instead, we build a quad-tree over the object mask area to cluster pixels with similar color as shown in~\cref{fig:image_sog} (see ~\cite{sog} for details).

\subsubsection{Object SoG}
\label{sec:gauss_to_sog}
We decode the canonical shape tokens $Z^{S\star}$ into a dense 3D Gaussian asset via SAM3D's Gaussian decoder.
Generally, these photorealistic assets are very dense with hundreds of thousands of Gaussians, which is unnecessary for tracking.
We therefore sparsify the decoded asset using farthest-point sampling and keep $2000$ Gaussians.
The resulting object representation is a set of 3D Gaussians $\mathcal{G}_{o} = \{(\mu_j,\Sigma_j,c_j)\}_{j=1}^{N_o}$ where $\mu_j\in\mathbb{R}^3$ is the mean, $\Sigma_j$ is the diagonal covariance, and $c_j$ is the color.
For faster convergence, we use an isotropic approximation by replacing each diagonal covariance with its average variance, following classical SoG tracking~\cite{sog}.
Although exact perspective projection generally yields an anisotropic 2D footprint, we approximate the projected footprint to be isotropic.
This enables the scalar closed-form Gaussian overlap in \cref{eq:gauss_overlap_iso} for a smooth image-alignment loss.
The anisotropic projection and its empirical trade-off are analyzed in Section 10.5 of the Supplementary Material.

Thus, SAM3D's dense Gaussians provide the canonical object geometry and appearance, while SoG provide the lightweight matching objective for tracking.

\subsubsection{SoG alignment term}
\label{sec:sog_alignment}
Now, we define the alignment energy as
\begin{equation}
E\!\left(\mathcal{K}_{I_t},\Psi_t(\mathcal{G}_o)\right)
=
\sum_{i\in \mathcal{K}_{I_t}}
\min\!\Big(
\sum_{j\in \Psi_t(\mathcal{G}_o)} \upsilon_{j,t}\, E_{ij},
\; E_{ii}
\Big).
\label{eq:sog_energy_visible_topdown2}
\end{equation} 
where $\Psi_t(\mathcal{G}_o)$ denotes the 2D object Gaussians obtained by projecting $\mathcal{G}_o$ into frame $t$ using the current object pose and camera.
The term $E_{ij}$ is the color-weighted overlap between an image Gaussian $i$ and a projected model Gaussian $j$, and $E_{ii}$ is the self-overlap of $i$ used for occlusion handling.
A key difference from Gaussian Splatting is the simplified occlusion handling, which allows computing Gaussian similarities without sorting and in parallel (sums) instead of sequential, back-to-front alpha blending.
Finally, $\upsilon_{j,t}\in\{0,1\}$ gates out object Gaussians that are occluded by the hand (\cref{sec:visibility_gate}).

The pairwise overlap for two 2D Gaussians $\mathcal{B}_i$ and $\mathcal{B}_j$ is
\begin{equation}
E_{ij} = d(c_i,c_j)\!\int_{\Omega} \mathcal{B}_i(u)\,\mathcal{B}_j(u)\,du,
\label{eq:eij_def}
\end{equation}
where $(c_i,c_j)$ are the colors, $\Omega\subset\mathbb{R}^2$ is the image plane and $d(c_i,c_j)$ is a color kernel (see Section 10.4 of the Supplementary Material for details).
With isotropic covariances, the overlap has closed form:
\begin{equation}
\int_{\Omega} \mathcal{B}_i(u)\mathcal{B}_j(u)\,du
=
2\pi\frac{\sigma_i^2\sigma_j^2}{\sigma_i^2+\sigma_j^2}
\exp\!\Big(-\frac{\|\mu_i-\mu_j\|^2}{2(\sigma_i^2+\sigma_j^2)}\Big).
\label{eq:gauss_overlap_iso}
\end{equation}

\begin{figure}[t]
  \centering
\includegraphics[trim=0 3 0 0, clip,width=1.0\linewidth]{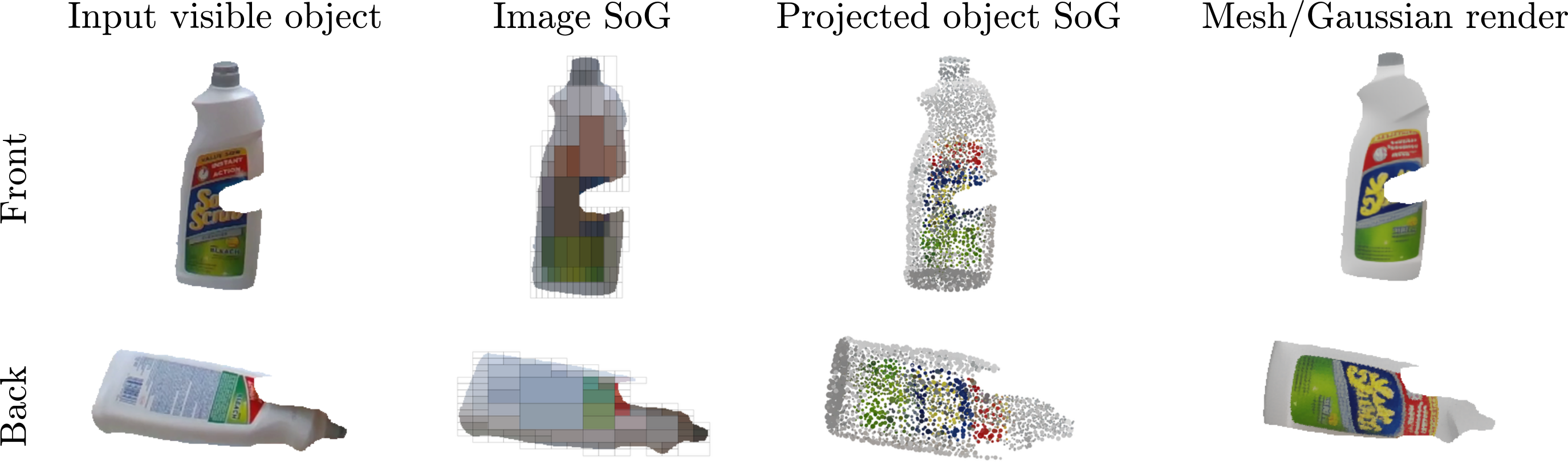}
   \caption{\textbf{Comparing mesh/Gaussians-based optimization to SoG.} 
   We show the differences between objectives in optimization using mesh/Gaussian rendering vs using SoG.
   Since MV-SAM3D/SAM3D has difficulty generating high-quality texture, specifically texture with text, trying to optimize the object pose with the rendered mesh/Gaussians does not work well, as shown in \cref{tab:ablation}.
   In contrast, the SoG-based tracking formulation matches image regions more approximately and thus, even with inaccurate texture, we can still optimize the object pose.
    }
   \label{fig:gaus_vs_sog}
\end{figure}

\subsubsection{Hand-aware SoG alignment}
\label{sec:visibility_gate}
Unlike full-body tracking~\cite{sog}, we leverage hand masks to restrict object-to-image matching to visible regions, preventing optimization drift during occlusions.
Specifically, we exclude occluded Gaussians from the alignment using the visibility gate $\upsilon_{j,t} = \mathbf{1}\!\left[M^h_t\!\left(\mathrm{round}(\mu^{2D}_{j,t})\right)=0\right]$
where $M^h_t$ is the binary hand mask and $\mu^{2D}_{j,t}$ is the projected mean of object Gaussian $j$.

\subsubsection{Geometric and interaction priors}
\label{sec:hand_losses}
The alignment term $E_t$ commits the projected object \emph{interior} to image evidence, but cannot, on its own, resolve hand articulation, metric depth, or contact under occlusion. We therefore couple it with a small set of priors that bias the optimizer toward physically and temporally consistent solutions.

We pull projected hand and object boundaries\linebreak toward the predicted-mask contours through a 2D signed distance-transform field,
as shown in \cref{fig:stage3}. Let $\Phi^x_t$ denote the signed distance transform of mask $M^x_t$, with $x\in\{o,h\}$, and let $\mathcal{B}^x_t$ be the visible boundary of the rendered shape (MANO silhouette vertices for the hand; projected object Gaussians on the rendered alpha contour for the object). Then, the silhouette loss can be defined as:
\begin{equation}
\mathcal{L}_{\text{sil}} = \sum_{t\in\mathcal{W}}\sum_{x\in\{o,h\}} \frac{1}{|\mathcal{B}^x_t|} \sum_{u\in\mathcal{B}^x_t} \big| \Phi^x_t(u) \big|
\label{eq:sil}
\end{equation}
This boundary-only loss yields dense gradients for pose and articulation without requiring asset's photometric quality.

Dyn-HaMR~\cite{yu2025dyn} already provides strong temporal hand tracking and predicts per-frame hand poses $\{\boldsymbol{\Theta}_t, R^h_t, \tau^h_t\}_{t=1}^{N}$, where $\boldsymbol{\Theta}_t$ denotes articulated joint pose parameters, $R^h_t$ the global root orientation, and $\tau^h_t$ the global translation.
Therefore, we do not employ SoG tracking for the hand; instead, we regularize the initialized MANO trajectory with lightweight geometric priors: 2D joint reprojection, depth consistency, and local contact.
We denote the joint reprojection term by $\mathcal{L}_{\text{j2d}}$ and provide its full formulation in Section 10.4 of the Supplementary Material
To resolve depth ambiguity while staying robust to noisy monocular depth near occlusion boundaries, we use the pointmap $P$ as an aggregate depth anchor rather than a dense objective.
Let $D_t^{o}$ and $D_t^{h}$ denote rendered depths for the object and hand, and let $D_t^{\mathrm{pm}}$ denote the pointmap depth.
We summarize these maps by median depths in the object and hand masks:
$z_t^x=\operatorname{median}(D_t^x[M_t^x])$ and
$z_t^{\mathrm{pm},x}=\operatorname{median}(D_t^{\mathrm{pm}}[M_t^x])$ for $x\in\{o,h\}$.
Our depth term is
\begin{equation}
\begin{split}
\mathcal{L}_{\text{depth}} &=
\sum_{t\in\mathcal{W}}
\left(
\left\|z_t^h-z_t^{\mathrm{pm},h}\right\|_2^2
+
\left\|z_t^o-z_t^{\mathrm{pm},o}\right\|_2^2
+
c_t^{ho}\left\|z_t^o-z_t^h\right\|_2^2
\right)
\end{split}
\label{eq:depth_prior}
\end{equation}
The first two terms anchor hand and object to the pointmap's metric scale.
The third term is active in contact frames and ties the object depth to the hand, reducing failures where the object pointmap alone gives a visually plausible but grasp-inconsistent depth.

To ensure physical plausibility, we introduce a surface-aware contact loss $\mathcal{L}_{\text{contact}}$ that encourages designated hand contact zones~\cite{hasson2019learning} to align with the object surface while penalizing interpenetration.
We additionally use a smoothness loss $\mathcal{L}_{\text{smooth}}$ that penalizes temporal acceleration in translation and rotation.
Detailed mathematical formulations of these tracking priors, along with their associated hyperparameters and weights in \cref{eq:final_obj}, are provided in Sections 10.2 and 10.4 of the Supplementary Material.

This optimization is highly lightweight because both the image and the object are processed as compact SoGs: the image is summarized by a quad-tree of 2D Gaussians, and the object is downsampled to at most 2000 Gaussians via farthest point sampling.
This design revives the efficiency of classical SoG tracking~\cite{sog}, while leveraging generative 3D models to obtain an accurate canonical object and a strong initialization in the wild.

\begin{figure*}[h]
  \centering
   \includegraphics[width=0.90\linewidth]{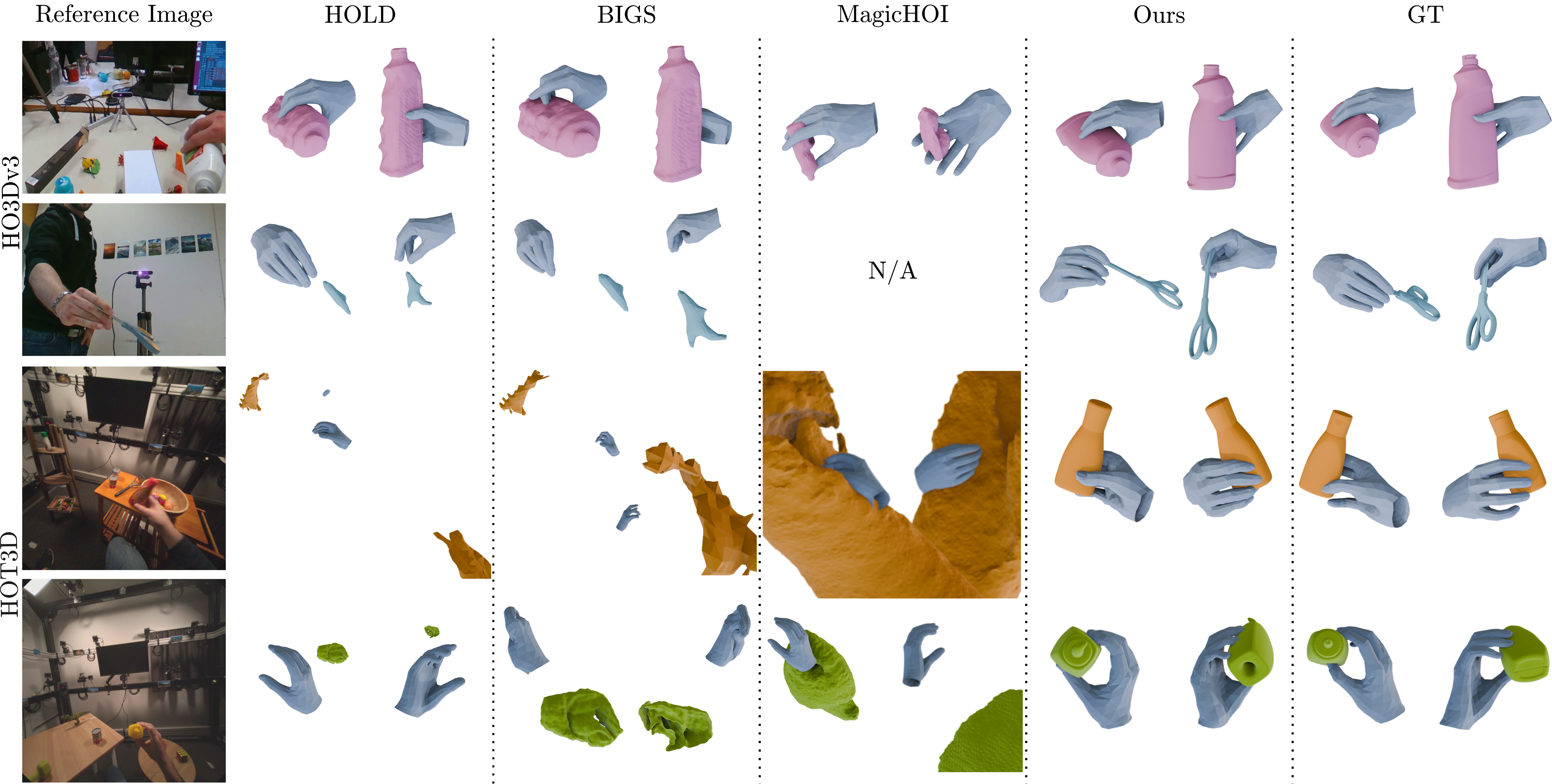}
   \caption{\textbf{Qualitative Comparison.}
   We compare the output of GraG with previous SoTA works HOLD, BIGS, and MagicHOI on HO3D (first 2 rows) and HOT3D (last 2 rows). 
   In the 2nd row 4th column, MagicHOI fails to produce a valid reconstruction; we therefore report it as \texttt{N/A}.
   Overall, GraG preserves sharper object geometry and yields more plausible hand poses (with fewer interpenetrations), while being substantially more efficient.
   }
   \label{fig:qualitative_results}
\end{figure*}

\section{Experiments}
\label{sec:experiments}

\subsection{Datasets}
\label{subsec:datasets}
\paragraph{HO3Dv3 Dataset}
HO3Dv3~\cite{hampali2021ho3dv3} contains monocular videos of single-hand object interactions.
We use the 18 sequences provided in the HOLD~\cite{fan2024hold} evaluation set.

\paragraph{HOT3D}
HOT3D~\cite{banerjee2025hot3d} is an egocentric dataset with accurate 3D poses and shapes of hands and objects.
Since the dataset does not provide ground-truth annotations for its test set, we instead select 18 sequences from the training set. 
Specifically, we extract segments from the full training videos that contain only single hand–object interactions with different objects.
Before running the methods on HOT3D, the fisheye videos are undistorted.

\subsection{Metrics}
\label{subsec:metrics}
We decode shape tokens into meshes via SAM3D's decoder and evaluate following HOLD~\cite{fan2024hold}. 
Hand accuracy is measured via root-relative MPJPE (mm). 
To evaluate object template quality independent of pose, we ICP-align meshes to compute Chamfer distance (CD, cm) and F-score at 10~mm (F10)~\cite{fan2024hold}.
To measure object pose and shape relative to the hand in 3D, we subtract the predicted hand root from the object vertices and compute the hand-relative CD for the object (CD$_h$)~\cite{fan2024hold}.
We also report Success Rate (SR) where a sequence is deemed as a failure if the method fails to output a valid result such that when initialization breaks down (notably for SfM-based pipelines) or when tracking degrades severely such that the hand-relative CD reaches 1000 cm.
As methods operate with different clip lengths, we report normalized runtime (hours per 100 frames) on a single NVIDIA RTX4090 GPU to quantify efficiency, including preprocessing and optimization.

\subsection{Implementation Details}
\label{subsec:implementation}
We optimize each sequence with AdamW~\cite{adamw} in a sliding-window manner:  running 100 iterations per window of 8 frames (stride 7), sequentially covering the full video.
Within each window, we optimize the hand and object pose parameters, while keeping the canonical object and hand shape fixed after initialization.
On a single RTX4090 GPU, for a 100-frame sequence, Stages 1-2 take $\sim$6 minutes on average and Stage~3 takes $\sim$10 minutes.
See Section 10.2 of the Supplementary Material for the hyper-parameters.

\begin{figure*}[t]
  \centering
\includegraphics[trim={0, 5, 0, 0}, clip, width=1.0\linewidth]{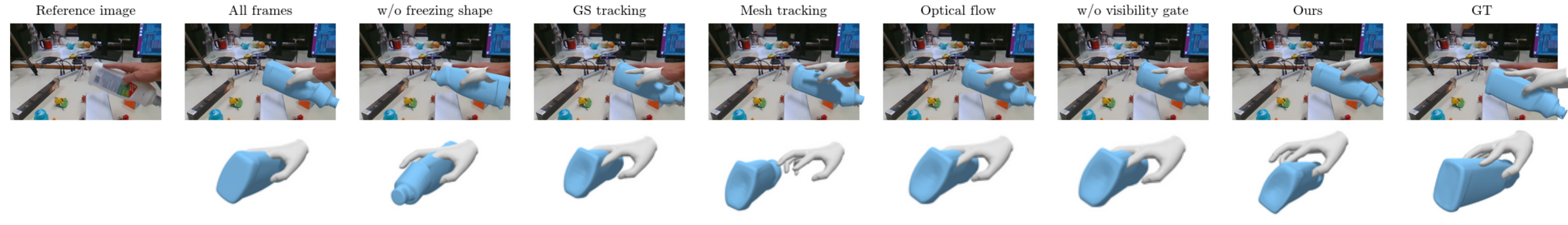}
   \caption{\textbf{Ablation experiments.} 
   We visualize how key design choices affect reconstruction quality (two representative views per setting: camera view and back view).
Using all frames during reconstruction yields an inaccurate canonical object shape and scale, leading to implausible grasps.
Running Stage~2 SAM3D pose estimation without freezing the shape tokens destabilizes per-frame pose estimates (note that Stage~3 still utilizes the Stage~1 canonical object in this experiment).
Replacing our compact SoG refinement with dense Gaussian Splatting (GS) tracking is slower and often fails to converge under the same iteration budget.
Similarly, replacing SoG with optical flow causes tracking to collapse during hand-object occlusion, degrading the final reconstruction.
   Without the visibility gate, SoG alignment is biased by hand-occluded regions, producing incorrect object poses.
Our full model achieves the highest fidelity, closely matching the ground truth.
}
\label{fig:fig_ablation}
\end{figure*}

\subsection{Results}
\label{subsec:results}
\paragraph{Compared SoTA Methods}
We compare GraG to state-of-the-art methods including HOLD~\cite{fan2024hold}, BIGS~\cite{on2025bigs}, and MagicHOI~\cite{wang2025magichoi}, following HOLD's evaluation protocol.
HOLD is a NeRF-based method that jointly optimizes hand–object scenes using SDF representations.
BIGS reconstructs HOI scenes using 3D Gaussians with a triplane MLP and SDS loss.
MagicHOI extends HOLD by leveraging novel-view synthesis models to hallucinate occluded object regions.
Note that while MagicHOI is designed for short clips specifically, we still include the comparison for the sake of completeness.
These are full-system comparisons: GraG benefits from stronger object and hand priors (SAM3D, Dyn-HaMR) that were not available in HOLD or BIGS. The quantitative gains reflect both these priors and our lightweight tracking formulation.

\paragraph{Quantitative Results}
We report quantitative results in \cref{tab:quant_res}.
On HO3Dv3, our method achieves the best object reconstruction quality, reducing $CD$ by 23.9\% over BIGS and 34.6\% over HOLD while increasing F10 to 97.2\%, partly due to the stronger SAM3D-based object prior.
GraG also obtains the second-best MPJPE (9.38\,mm) and preserves a 100\% success rate.
MagicHOI obtains the lowest MPJPE on successful clips, but its lower success rate and high $CD_h$ indicate unstable hand-object alignment on full sequences.
Notably, GraG is also substantially faster, requiring 0.27 hours per 100 frames compared to 1.2 for MagicHOI, 3.6 for BIGS, and 10.5 for HOLD. 

\begin{table}[t]
\centering
\caption{
\textbf{Quantitative comparison.}
We report object reconstruction, hand accuracy, success rate, and runtime on HO3Dv3 and HOT3D.
We highlight the \colorbox{pastelgreendark}{best} and \colorbox{pastelgreen}{second-best} results.
All methods are averaged over the \textit{successful results} only. 
\textsuperscript{(*)}BIGS results on HO3Dv3 differ from those reported by~\cite{on2025bigs}; we obtain them by running the authors' open-source implementation with the reported hyperparameters.
}
\label{tab:quant_res}
\centering
\small
\resizebox{\linewidth}{!}{%
\begin{tabular}{llcccccc}
\toprule
Dataset & Method 
& \shortstack{CD \\ $[\mathrm{cm}]$} $\downarrow$
& \shortstack{F10 \\ $[\%]$} $\uparrow$
& \shortstack{MPJPE \\ $[\mathrm{mm}]$} $\downarrow$
& \shortstack{$\mathrm{CD}_h$ \\ $[\mathrm{cm}]$} $\downarrow$
& \shortstack{Success Rate \\ $[\%]$} $\uparrow$
& \shortstack{Runtime \\ $[\mathrm{h}]$} $\downarrow$ \\
\midrule

\multirow{4}{*}{HO3Dv3}
& HOLD
& 0.78
& 92.0
& 23.4
& \best{\textbf{4.27}}
& \best{\textbf{100}}
& 10.5 \\

& BIGS\textsuperscript{*}
& \second{0.67}
& \second{94.1}
& 23.9
& 7.61
& \best{\textbf{100}}
& 3.60 \\

& MagicHOI
& 1.58
& 73.7
& \best{\textbf{4.35}}
& 129
& \second{67.0}
& \second{1.20} \\

& \textbf{Ours}
& \best{\textbf{0.51}}& \best{\textbf{97.2}}& \second{9.38}& \second{5.02}& \best{\textbf{100}}
& \best{\textbf{0.27}}\\

\midrule

\multirow{4}{*}{HOT3D}
& HOLD
& 3.14
& \second{53.3}
& 21.9
& 122
& 50.0
& 10.5 \\

& BIGS
& \second{2.92}
& 51.7
& 36.9
& \second{84.2}
& 50.0
& 3.60 \\

& MagicHOI
& 22.6
& 33.6
& \best{\textbf{18.4}}
& 97.1
& \second{55.6}
& \second{1.20} \\

& \textbf{Ours}
& \best{\textbf{0.67}}& \best{\textbf{94.7}}& \second{20.6}& \best{\textbf{9.37}}& \best{\textbf{100}}
& \best{\textbf{0.27}}\\

\bottomrule
\end{tabular}%
}
\end{table}

On the more challenging HOT3D dataset, characterized by egocentric viewpoints, rapid camera motion, and heavy hand occlusions, our method further improves both object reconstruction and hand–object alignment.
We reduce $CD$ by 77.1\% over BIGS and 78.7\% over HOLD, and achieve the best hand-object alignment, reducing $CD_h$ by 88.9\% over the strongest baseline.
Our method also achieves a 100\% success rate, whereas SfM-based baselines drop to 50–56\%.
In long egocentric sequences, COLMAP-based initialization is sensitive to frame sampling and often produces unstable camera poses, which leads to drift or failed reconstructions.
Consequently, methods that depend on such initialization (HOLD, BIGS, and MagicHOI) frequently fail to converge or produce inconsistent object tracking.
Specifically, MagicHOI is designed for short-clip evaluation; when applied to our full-length HO3Dv3 and egocentric HOT3D sequences, its COLMAP-based initialization frequently becomes unstable, reducing its success rate.

\paragraph{Qualitative Results}
As shown in Fig. \ref{fig:qualitative_results}, our method accurately reconstructs hands and objects from video, preserving finer geometric details and more coherent spatial placement than prior SoTA methods. 
Specifically on HOT3D dataset, our approach demonstrates superior performance; it recovers thin structures (e.g., scissors) and maintains plausible object scale under occlusion, whereas competing methods either fail to reconstruct meaningful object geometry or diverge during optimization, largely due to their reliance on COLMAP/SfM initialization.
Our results exhibit fewer misalignments and more realistic grasping configurations. See Fig. 13 in the Supplementary Material for overlays on reference images.

\section{Ablations}
\label{sec:ablations}
We validate our design choices by ablating on two HO3Dv3 sequences and two HOT3D sequences.
\begin{table}[t]
\caption{\textbf{Ablation study.} We ablate key components of our method on 4 sequences, 2 sequences from HO3Dv3 and 2 sequences from HOT3D.
The first block ablates Stage~1 object reconstruction, the second block ablates Stage~2 SAM3D pose adaptation, and the third block ablates Stage~3 tracking objectives.
For Stage~2 and Stage~3 variants, the same canonical object reconstructed with balanced keyframes is used, so CD and F10 remain unchanged.
The ``w/o freezing shape'' row does not freeze the shape latent during SAM3D pose adaptation, but still uses the Stage~1 object during Stage~3 tracking.
Runtime is normalized per 100 frames.
}
\centering
\resizebox{\columnwidth}{!}{%
\begin{tabular}{lccccc}
\toprule
{Ablation}
& \shortstack{CD \\ $[\mathrm{cm}]\downarrow$} 
& \shortstack{F10 \\ $[\%]\uparrow$}
& \shortstack{MPJPE \\ $[\mathrm{mm}]\downarrow$}
& \shortstack{$\mathrm{CD}_h$ \\ $[\mathrm{cm}]\downarrow$} 
& \shortstack{Runtime \\ $[\mathrm{h}]\downarrow$} \\
\midrule
random $K=4$ keyframe selection & 0.69 & 93.9 & 11.5 & 9.36& 0.27\\
single view from first frame & 0.73 & 92.1 & 18.2 & 22.73 & 0.27\\
using all frames& 0.68 & 93.6 & 11.5 & 9.95&0.30\\
\midrule
w/o freezing shape & 0.45& 98.8 & 11.4 & 10.81 & 0.28 \\
w/o temporal guidance ($\lambda_{\text{temp}}{=}0$)  & 0.45& 98.8 & 11.5 & 10.60 & 0.27\\
\midrule
only foundation models (w/o Stage 3)                             & 0.45& 98.8 & 12.1& 20.62 & \best{\textbf{0.10}}\\
Gaussian Splatting tracking instead of SoG           & 0.45& 98.8 & 11.8&  10.46& 1.20\\
 mesh rendering instead of SoG& 0.45& 98.8 & 13.6 & 12.60 &2.40 \\
 optical flow tracking instead of SoG& 0.45& 98.8 & 11.8& 11.45&0.24 \\
opt. w/o SoG refinement (only prior losses)& 0.45& 98.8 & 11.5& 10.45& 0.21 \\
w/o visibility gate (no hand-occlusion gating)       & 0.45& 98.8 & 11.5 & 9.35 & 0.27\\
w/o depth loss $\mathcal{L}_{\text{depth}}$         & 0.45& 98.8 & 11.4& 12.70& 0.27\\
w/o silhouette loss $\mathcal{L}_{\text{sil}}$       & 0.45 & 98.8 & 11.5 & 10.71& 0.25\\
w/o contact loss $\mathcal{L}_{\text{contact}}$       & 0.45 & 98.8 & 11.5 & 19.58 & 0.21 \\
\midrule
\textbf{full model}                                  & \best{\textbf{0.45}} & \best{\textbf{98.8}} & \best{\textbf{11.3}} & \best{\textbf{7.65}} & 0.27\\
\bottomrule
\end{tabular}%
}
\label{tab:ablation}
\end{table}

\begin{figure*}[t]
    \centering

    \begin{minipage}[t]{0.48\textwidth}
        \centering
        \includegraphics[height=7.2cm,keepaspectratio]{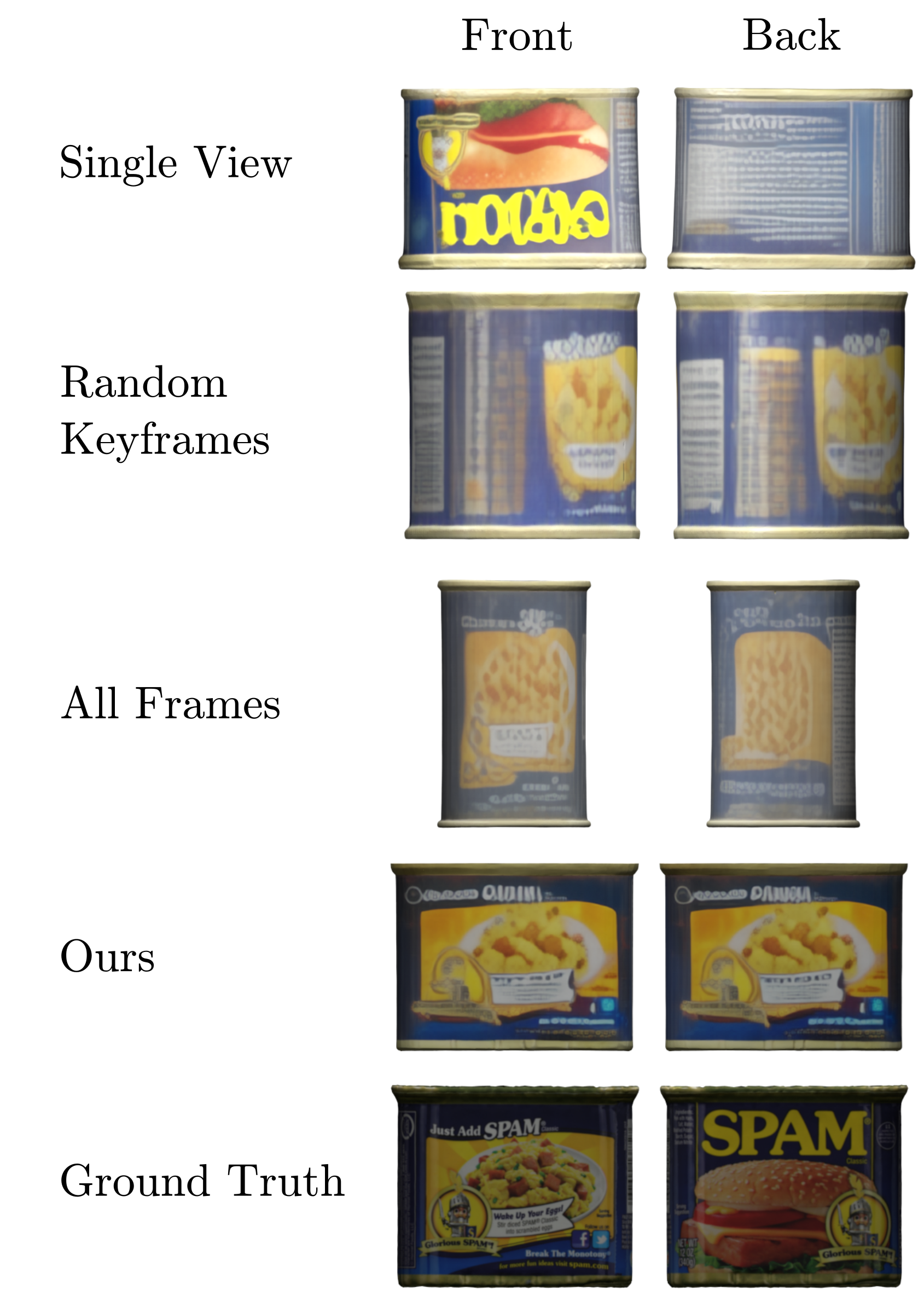}
        \caption{\textbf{Effect of keyframe selection on object reconstruction.}
        Random keyframe selection and using all frames for reconstruction degrade
        the canonical reconstruction (shape/scale) compared to our balanced
        keyframe selection; ground truth shown for reference.}
        \label{fig:keyframe_ablation}
    \end{minipage}
    \hfill
    \begin{minipage}[t]{0.48\textwidth}
        \centering
        \includegraphics[width=\linewidth]{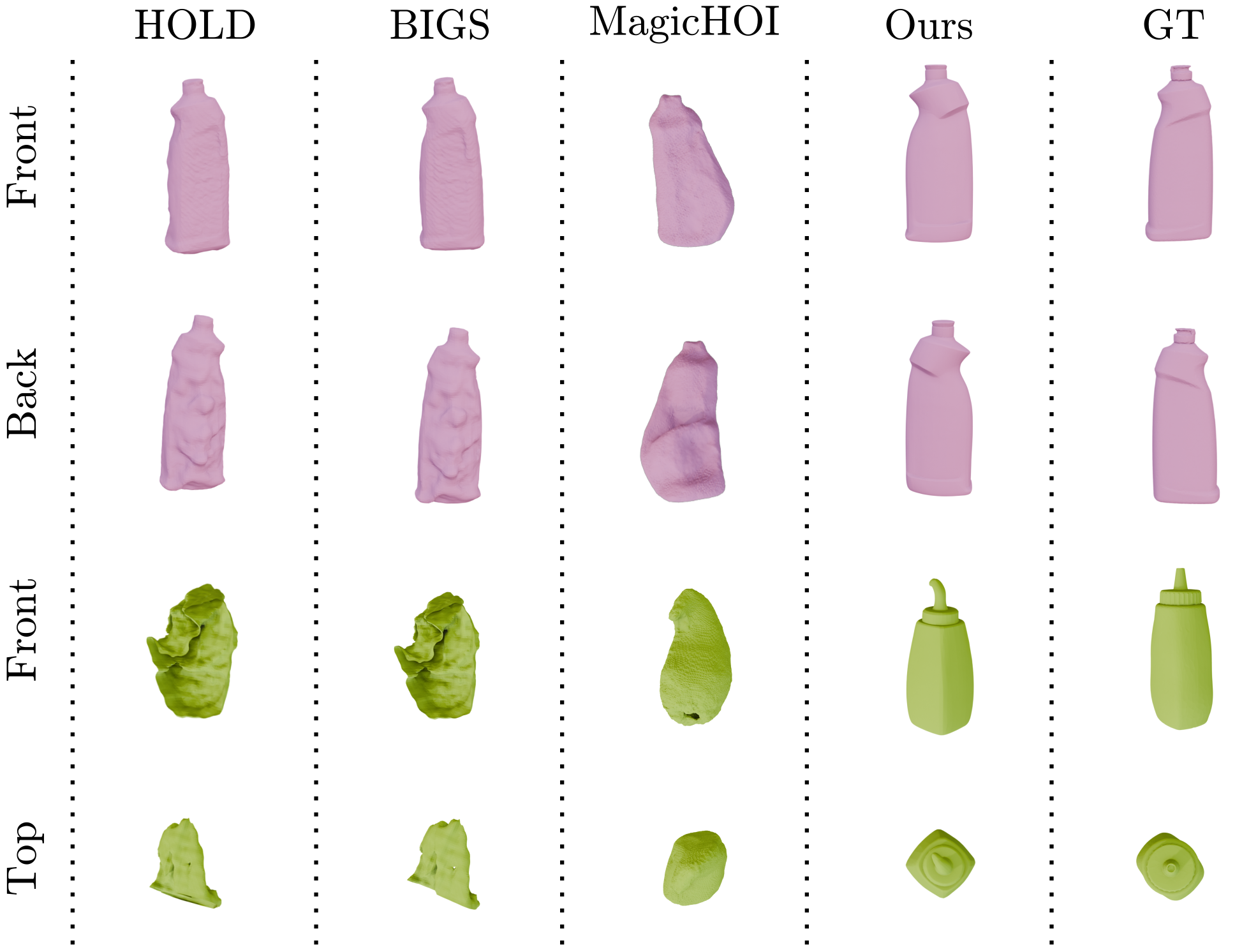}
        \caption{\textbf{Object-only reconstruction comparison.}
        Front/back views of reconstructed object geometry; ground truth shown
        for reference.}
        \label{fig:obj_only}
    \end{minipage}

\end{figure*}

\begin{figure*}[h]
  \centering
  \includegraphics[width=1.0\textwidth]{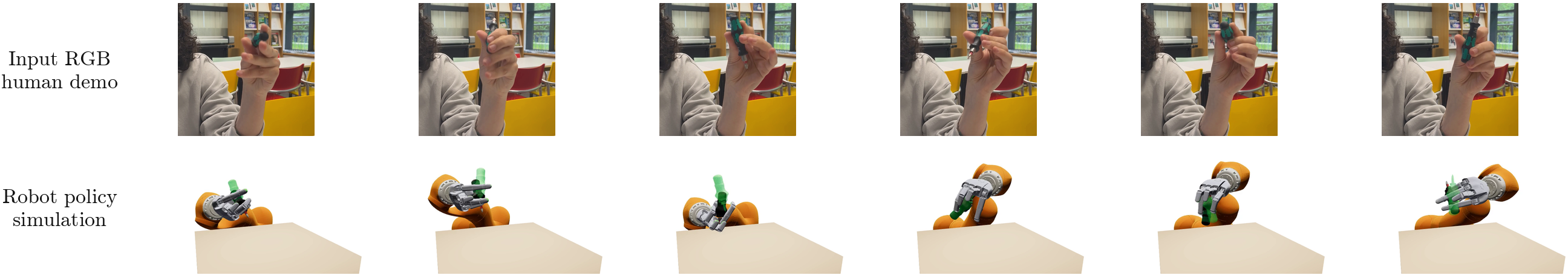}
  \caption{\textbf{Human-to-robot tool manipulation from GraG object trajectories.}
  Top: monocular RGB frames from a human short-screwdriver demonstration. Bottom: corresponding visualizations of the
SimToolReal policy observations. The green screwdriver indicates the current goal pose obtained from GraG's
reconstructed 6D object trajectory (Zoom in for best view). The policy receives the robot state, estimated object pose, and goal pose; once the current
object pose is sufficiently close to the goal, the goal advances to the next pose in the sequence. This visualization renders the
policy inputs and goal-tracking behavior, rather than a physics-based real-robot execution. 
  }
  \label{fig:robot_sim}
\end{figure*}

Results are presented in~\cref{tab:ablation}, ~\cref{fig:fig_ablation} and the Supplementary Material.
For Stage~1, balanced keyframes give better object reconstruction than single frame, random keyframes or using all frames, showing that view quality is more important than simply increasing the number of SAM3D inputs, as also shown in \cref{fig:keyframe_ablation}.
In Stage~2, freezing the canonical shape and adding temporal guidance improve hand-relative object pose error, indicating more stable video-level pose adaptation.
Most importantly, the ``only foundation models'' ablation evaluates the same SAM3D object and Dyn-HaMR hand priors without our Stage~3 refinement, and yields much higher hand-object error ($CD_h{=}20.62$).
The gap to the full model ($CD_h{=}7.65$) comes from the whole Stage~3, which uses both the pretrained priors and the SoG alignment, and should not be read as the effect of SoG alone.
When we drop only the SoG tracking term but keep the Stage~3 priors, $CD_h$ is already $10.45$, which shows that the priors recover most of the error after ``only foundation models'', and SoG closes the remaining gap (from $10.45$ to $7.65$).
The priors are what make the hand-relative pose accurate, while the compact hand-aware SoG mainly provides an efficient tracking signal.
We further replace SoG with dense Gaussian rendering, mesh rendering, and Lucas--Kanade\cite{lucas1981iterative} optical-flow tracking baselines.
Dense Gaussian and mesh rendering optimize RGB render-and-compare losses from the SAM3D asset, but are sensitive to generated texture and hand occlusion.
Optical flow tracks projected object surface points through the video, but tracks are often lost under occlusion, fast rotation, and large appearance changes.
All alternatives underperform the full SoG objective, mesh and dense Gaussian tracking is also slower on the ablation set.
Among losses, contact is the most critical for hand-object alignment; depth, silhouette, and visibility gating improve robustness to occlusion.

\section{Application}
To illustrate a downstream robotic use case, we interface GraG with SimToolReal~\cite{simtoolreal}, an object-centric framework for zero-shot dexterous tool manipulation. 
SimToolReal represents a task as a sequence of desired 6D object goal poses, and its policy observes the current robot/object state together with the relative object-to-goal keypoints. 
We record a monocular human demonstration of an in-house short screwdriver, reconstruct its 6D object trajectory with GraG, and convert this trajectory into the world-frame goal-pose sequence used by SimToolReal. 
We then evaluate the pretrained short-screwdriver policy on the corresponding vertical-spinning task, and visualize it in \cref{fig:robot_sim} and the supplementary video. 
As shown in the supplementary video, SimToolReal successfully manipulates the screwdriver based on the object geometry and poses specified by GraG.  
It is noteworthy that while robust, incorrect object tracking contributes to as many as $44\%$ failure cases in  SimToolReal.
Although demonstrated qualitatively, GraG's ability to effortlessly integrate with SimToolReal demonstrates the generalization potential of our method.

\section{Limitations}
While using Depth Anything 3 pointmaps as aggregate metric anchors minimizes depth error, severe pointmap failures can still bias object scale and translation.
Similarly, our color-weighted SoG alignment is robust to texture imperfections but remains vulnerable to illumination changes, shadows, specularities, and weak object texturing.
Our alignment term approximately projects 3D primitives to isotropic 2D footprints.
This can lose fidelity under large viewpoint changes or partial cropping near the image boundary, where exact perspective projection may produce elliptical 2D footprints.
We use this approximation to have a simple but closed-form scalar overlap. Extending both object and image SoGs to anisotropic 2D covariances or a hybrid version would be an interesting future direction.
Our pipeline also depends on high-quality hand and object masks, which can be challenging to recover under extreme occlusion.
Finally, the current system is restricted to single-hand manipulation of a single rigid object.
We leave bimanual interactions and deformable objects for the future work to explore.

\section{Conclusion}
\label{sec:conclusion}
We introduced Grasp in Gaussians (\textbf{GraG}), an efficient and scalable framework for reconstructing dynamic 3D hand-object interactions from a single monocular video.
We showed that na\"ively combining the 2D/3D foundational models can produce suboptimal results, particularly in terms of optimization runtime.
We addressed this by re-introducing the idea of Sum-of-Gaussian tracking and showed how a sparse set of Gaussians can be tracked efficiently.
Experiments demonstrate that, as a full system built on modern object and hand priors, GraG improves reconstruction quality and robustness while substantially reducing runtime compared to prior video-based HOI reconstruction methods.
In future work, we aim to integrate physical plausibility of the reconstruction to improve robustness and extend the pipeline to multiple hands and interacting objects.

\begin{acks}
This project was supported by the Saarbrücken Research Center for Visual Computing, Interaction, and AI. We would like to thank the anonymous reviewers for constructive comments and suggestions.
\end{acks}

\bibliographystyle{ACM-Reference-Format}
\bibliography{main}

@String(PAMI  = {IEEE Trans. Pattern Anal. Mach. Intell.})

@String(CVPR  = {IEEE Conf. Comput. Vis. Pattern Recog.})

@String(IJCAI = {IJCAI})

@String(TOG   = {ACM Trans. Graph.})

@String(PAMI  = {IEEE TPAMI})

@String(CVPR  = {CVPR})

@String(TOG   = {ACM TOG})

@inproceedings{fan2024hold,
  title={Hold: Category-agnostic 3d reconstruction of interacting hands and objects from video},
  author={Fan, Zicong and Parelli, Maria and Kadoglou, Maria Eleni and Chen, Xu and Kocabas, Muhammed and Black, Michael J and Hilliges, Otmar},
  booktitle={Proceedings of the IEEE/CVF Conference on Computer Vision and Pattern Recognition},
  pages={494--504},
  year={2024}
}

@inproceedings{on2025bigs,
  title={BIGS: Bimanual Category-agnostic Interaction Reconstruction from Monocular Videos via 3D Gaussian Splatting},
  author={On, Jeongwan and Gwak, Kyeonghwan and Kang, Gunyoung and Cha, Junuk and Hwang, Soohyun and Hwang, Hyein and Baek, Seungryul},
  booktitle={Proceedings of the Computer Vision and Pattern Recognition Conference},
  pages={17437--17447},
  year={2025}
}

@inproceedings{liu2025easyhoi,
  title={EasyHOI: Unleashing the Power of Large Models for Reconstructing Hand-Object Interactions in the Wild},
  author={Liu, Yumeng and Long, Xiaoxiao and Yang, Zemin and Liu, Yuan and Habermann, Marc and Theobalt, Christian and Ma, Yuexin and Wang, Wenping},
  booktitle={Proceedings of the Computer Vision and Pattern Recognition Conference},
  pages={7037--7047},
  year={2025}
}

@article{wang2025magichoi,
  title={MagicHOI: Leveraging 3D Priors for Accurate Hand-object Reconstruction from Short Monocular Video Clips},
  author={Wang, Shibo and He, Haonan and Parelli, Maria and Gebhardt, Christoph and Fan, Zicong and Song, Jie},
  journal={arXiv preprint arXiv:2508.05506},
  year={2025}
}

@INPROCEEDINGS{aytekin2025follow,

  author={Aytekin, Ayce Idil and Rhodin, Helge and Dabral, Rishabh and Theobalt, Christian},

  booktitle={2026 International Conference on 3D Vision (3DV)}, 

  title={Follow My Hold: Hand-Object Interaction Reconstruction Through Geometric Guidance}, 

  year={2026},

  volume={},

  number={},

  pages={795-805},

  doi={10.1109/3DV69130.2026.00081}
}

@inproceedings{hasson2020leveraging,
  title={Leveraging photometric consistency over time for sparsely supervised hand-object reconstruction},
  author={Hasson, Yana and Tekin, Bugra and Bogo, Federica and Laptev, Ivan and Pollefeys, Marc and Schmid, Cordelia},
  booktitle={Proceedings of the IEEE/CVF conference on computer vision and pattern recognition},
  pages={571--580},
  year={2020}
}

@inproceedings{ye2022s,
  title={What's in your hands? 3d reconstruction of generic objects in hands},
  author={Ye, Yufei and Gupta, Abhinav and Tulsiani, Shubham},
  booktitle={Proceedings of the IEEE/CVF conference on computer vision and pattern recognition},
  pages={3895--3905},
  year={2022}
}

@inproceedings{ye2023diffusion,
  title={Diffusion-guided reconstruction of everyday hand-object interaction clips},
  author={Ye, Yufei and Hebbar, Poorvi and Gupta, Abhinav and Tulsiani, Shubham},
  booktitle={Proceedings of the IEEE/CVF international conference on computer vision},
  pages={19717--19728},
  year={2023}
}

@inproceedings{pavlakos2024hamer,
  title={Reconstructing hands in 3d with transformers},
  author={Pavlakos, Georgios and Shan, Dandan and Radosavovic, Ilija and Kanazawa, Angjoo and Fouhey, David and Malik, Jitendra},
  booktitle={Proceedings of the IEEE/CVF Conference on Computer Vision and Pattern Recognition},
  pages={9826--9836},
  year={2024}
}

@inproceedings{yu2025dyn,
  title={Dyn-hamr: Recovering 4d interacting hand motion from a dynamic camera},
  author={Yu, Zhengdi and Zafeiriou, Stefanos and Birdal, Tolga},
  booktitle={Proceedings of the Computer Vision and Pattern Recognition Conference},
  pages={27716--27726},
  year={2025}
}

@InProceedings{xiang2025trellis,
    author    = {Xiang, Jianfeng and Lv, Zelong and Xu, Sicheng and Deng, Yu and Wang, Ruicheng and Zhang, Bowen and Chen, Dong and Tong, Xin and Yang, Jiaolong},
    title     = {Structured 3D Latents for Scalable and Versatile 3D Generation},
    booktitle = {Proceedings of the IEEE/CVF Conference on Computer Vision and Pattern Recognition (CVPR)},
    month     = {June},
    year      = {2025},
    pages     = {21469-21480}
}

@BOOK{hartley2011mvgeo,
  title     = "Multiple View Geometry in Computer Vision",
  author    = "Hartley, Richard and Zisserman, Andrew",
  publisher = "Cambridge University Press",
  edition   =  2,
  month     =  jan,
  year      =  2011,
  address   = "Cambridge, England"
}

@ARTICLE{tomasi1992shape,
  title     = "Shape and motion from image streams under orthography: a
               factorization method",
  author    = "Tomasi, Carlo and Kanade, Takeo",
  journal   = "Int. J. Comput. Vis.",
  publisher = "Springer Science and Business Media LLC",
  volume    =  9,
  number    =  2,
  pages     = "137--154",
  month     =  nov,
  year      =  1992,
  language  = "en"
}

@INPROCEEDINGS{scharstein2002cw,
  title      = "A taxonomy and evaluation of dense two-frame stereo
                correspondence algorithms",
  booktitle  = "Proceedings {IEEE} Workshop on Stereo and {Multi-Baseline}
                Vision ({SMBV} 2001)",
  author     = "Scharstein, D and Szeliski, R and Zabih, R",
  publisher  = "IEEE Comput. Soc",
  year       =  2002,
  conference = "IEEE Workshop on Stereo and Multi-Baseline Vision (SMBV 2001)",
  location   = "Kauai, HI, USA"
}

@BOOK{szeliski2010sw,
  title     = "Computer Vision",
  author    = "Szeliski, Richard",
  publisher = "Springer",
  series    = "Texts in computer science",
  month     =  oct,
  year      =  2010,
  address   = "London, England",
  copyright = "https://www.springernature.com/gp/researchers/text-and-data-mining",
  language  = "en"
}

@inproceedings{deitke2023objaverse,
  title={Objaverse: A universe of annotated 3d objects},
  author={Deitke, Matt and Schwenk, Dustin and Salvador, Jordi and Weihs, Luca and Michel, Oscar and VanderBilt, Eli and Schmidt, Ludwig and Ehsani, Kiana and Kembhavi, Aniruddha and Farhadi, Ali},
  booktitle={Proceedings of the IEEE/CVF conference on computer vision and pattern recognition},
  pages={13142--13153},
  year={2023}
}

@inproceedings{sarlin2019coarse,
  title={From coarse to fine: Robust hierarchical localization at large scale},
  author={Sarlin, Paul-Edouard and Cadena, Cesar and Siegwart, Roland and Dymczyk, Marcin},
  booktitle={Proceedings of the IEEE/CVF conference on computer vision and pattern recognition},
  pages={12716--12725},
  year={2019}
}

@inproceedings{sarlin2020superglue,
  title={Superglue: Learning feature matching with graph neural networks},
  author={Sarlin, Paul-Edouard and DeTone, Daniel and Malisiewicz, Tomasz and Rabinovich, Andrew},
  booktitle={Proceedings of the IEEE/CVF conference on computer vision and pattern recognition},
  pages={4938--4947},
  year={2020}
}

@inproceedings{cao2021reconstructing,
  title={Reconstructing hand-object interactions in the wild},
  author={Cao, Zhe and Radosavovic, Ilija and Kanazawa, Angjoo and Malik, Jitendra},
  booktitle={Proceedings of the IEEE/CVF international conference on computer vision},
  pages={12417--12426},
  year={2021}
}

@inproceedings{corona2020ganhand,
  title={Ganhand: Predicting human grasp affordances in multi-object scenes},
  author={Corona, Enric and Pumarola, Albert and Alenya, Guillem and Moreno-Noguer, Francesc and Rogez, Gr{\'e}gory},
  booktitle={Proceedings of the IEEE/CVF conference on computer vision and pattern recognition},
  pages={5031--5041},
  year={2020}
}

@inproceedings{liu2021semi,
  title={Semi-supervised 3d hand-object poses estimation with interactions in time},
  author={Liu, Shaowei and Jiang, Hanwen and Xu, Jiarui and Liu, Sifei and Wang, Xiaolong},
  booktitle={Proceedings of the IEEE/CVF conference on computer vision and pattern recognition},
  pages={14687--14697},
  year={2021}
}

@inproceedings{tekin2019h+o,
  title={H+ o: Unified egocentric recognition of 3d hand-object poses and interactions},
  author={Tekin, Bugra and Bogo, Federica and Pollefeys, Marc},
  booktitle={Proceedings of the IEEE/CVF conference on computer vision and pattern recognition},
  pages={4511--4520},
  year={2019}
}

@inproceedings{yang2021cpf,
  title={Cpf: Learning a contact potential field to model the hand-object interaction},
  author={Yang, Lixin and Zhan, Xinyu and Li, Kailin and Xu, Wenqiang and Li, Jiefeng and Lu, Cewu},
  booktitle={Proceedings of the IEEE/CVF international conference on computer vision},
  pages={11097--11106},
  year={2021}
}

@inproceedings{hasson2019learning,
  title={Learning joint reconstruction of hands and manipulated objects},
  author={Hasson, Yana and Varol, Gul and Tzionas, Dimitrios and Kalevatykh, Igor and Black, Michael J and Laptev, Ivan and Schmid, Cordelia},
  booktitle={Proceedings of the IEEE/CVF conference on computer vision and pattern recognition},
  pages={11807--11816},
  year={2019}
}

@inproceedings{huang2022reconstructing,
  title={Reconstructing hand-held objects from monocular video},
  author={Huang, Di and Ji, Xiaopeng and He, Xingyi and Sun, Jiaming and He, Tong and Shuai, Qing and Ouyang, Wanli and Zhou, Xiaowei},
  booktitle={SIGGRAPH Asia 2022 Conference Papers},
  pages={1--9},
  year={2022}
}

@inproceedings{fan2023arctic,
  title={ARCTIC: A dataset for dexterous bimanual hand-object manipulation},
  author={Fan, Zicong and Taheri, Omid and Tzionas, Dimitrios and Kocabas, Muhammed and Kaufmann, Manuel and Black, Michael J and Hilliges, Otmar},
  booktitle={Proceedings of the IEEE/CVF conference on computer vision and pattern recognition},
  pages={12943--12954},
  year={2023}
}

@inproceedings{hasson2021towards,
  title={Towards unconstrained joint hand-object reconstruction from rgb videos},
  author={Hasson, Yana and Varol, G{\"u}l and Schmid, Cordelia and Laptev, Ivan},
  booktitle={2021 International Conference on 3D Vision (3DV)},
  pages={659--668},
  year={2021},
  organization={IEEE}
}

@inproceedings{karunratanakul2020grasping,
  title={Grasping field: Learning implicit representations for human grasps},
  author={Karunratanakul, Korrawe and Yang, Jinlong and Zhang, Yan and Black, Michael J and Muandet, Krikamol and Tang, Siyu},
  booktitle={2020 International Conference on 3D Vision (3DV)},
  pages={333--344},
  year={2020},
  organization={IEEE}
}

@inproceedings{hampali2023hand,
  title={In-hand 3d object scanning from an rgb sequence},
  author={Hampali, Shreyas and Hodan, Tomas and Tran, Luan and Ma, Lingni and Keskin, Cem and Lepetit, Vincent},
  booktitle={Proceedings of the IEEE/CVF Conference on Computer Vision and Pattern Recognition},
  pages={17079--17088},
  year={2023}
}

@inproceedings{zhong2024color,
  title={Color-NeuS: Reconstructing neural implicit surfaces with color},
  author={Zhong, Licheng and Yang, Lixin and Li, Kailin and Zhen, Haoyu and Han, Mei and Lu, Cewu},
  booktitle={2024 International Conference on 3D Vision (3DV)},
  pages={631--640},
  year={2024},
  organization={IEEE}
}

@article{v2025glm,
  title={GLM-4.5 V and GLM-4.1 V-Thinking: Towards Versatile Multimodal Reasoning with Scalable Reinforcement Learning},
  author={V Team and Hong, Wenyi and Yu, Wenmeng and others},
  journal={arXiv preprint arXiv:2507.01006},
  year={2025}
}

@article{hampali2021ho3dv3,
  title={Ho-3d\_v3: Improving the accuracy of hand-object annotations of the ho-3d dataset},
  author={Hampali, Shreyas and Sarkar, Sayan Deb and Lepetit, Vincent},
  journal={arXiv preprint arXiv:2107.00887},
  year={2021}
}

@article{adamw,
  title={Decoupled weight decay regularization},
  author={Loshchilov, Ilya and Hutter, Frank},
  journal={arXiv preprint arXiv:1711.05101},
  year={2017}
}

@Article{kerbl3Dgaussians,
      author       = {Kerbl, Bernhard and Kopanas, Georgios and Leimk{\"u}hler, Thomas and Drettakis, George},
      title        = {3D Gaussian Splatting for Real-Time Radiance Field Rendering},
      journal      = {ACM Transactions on Graphics},
      number       = {4},
      volume       = {42},
      month        = {July},
      year         = {2023},
      url          = {https://repo-sam.inria.fr/fungraph/3d-gaussian-splatting/}
}

@inproceedings{zheng2025gaustar,
  title={GauSTAR: Gaussian Surface Tracking and Reconstruction},
  author={Zheng, Chengwei and Xue, Lixin and Zarate, Juan and Song, Jie},
  booktitle={Proceedings of the Computer Vision and Pattern Recognition Conference},
  pages={16543--16553},
  year={2025}
}

@inproceedings{sog,
  title={Fast articulated motion tracking using a sums of gaussians body model},
  author={Stoll, Carsten and Hasler, Nils and Gall, Juergen and Seidel, Hans-Peter and Theobalt, Christian},
  booktitle={2011 international conference on computer vision},
  pages={951--958},
  year={2011},
  organization={IEEE}
}

@article{sam3,
  title={Sam 3: Segment anything with concepts},
  author={Carion, Nicolas and Gustafson, Laura and Hu, Yuan-Ting and Debnath, Shoubhik and Hu, Ronghang and Suris, Didac and Ryali, Chaitanya and Alwala, Kalyan Vasudev and Khedr, Haitham and Huang, Andrew and others},
  journal={arXiv preprint arXiv:2511.16719},
  year={2025}
}

@article{sam3d,
  title={Sam 3d: 3dfy anything in images},
  author={Chen, Xingyu and Chu, Fu-Jen and Gleize, Pierre and Liang, Kevin J and Sax, Alexander and Tang, Hao and Wang, Weiyao and Guo, Michelle and Hardin, Thibaut and Li, Xiang and others},
  journal={arXiv preprint arXiv:2511.16624},
  year={2025}
}

@article{depthanything3,
  title={Depth Anything 3: Recovering the visual space from any views},
  author={Haotong Lin and Sili Chen and Jun Hao Liew and Donny Y. Chen and Zhenyu Li and Guang Shi and Jiashi Feng and Bingyi Kang},
  journal={arXiv preprint arXiv:2511.10647},
  year={2025}
}

@misc{mv-sam3d,
      title={MV-SAM3D: Adaptive Multi-View Fusion for Layout-Aware 3D Generation}, 
      author={Baicheng Li and Dong Wu and Jun Li and Shunkai Zhou and Zecui Zeng and Lusong Li and Hongbin Zha},
      year={2026},
      eprint={2603.11633},
      archivePrefix={arXiv},
      primaryClass={cs.CV},
      url={https://arxiv.org/abs/2603.11633}, 
}

@article{lai2025hunyuan3d,
  title={Hunyuan3D 2.5: Towards High-Fidelity 3D Assets Generation with Ultimate Details},
  author={Lai, Zeqiang and Zhao, Yunfei and Liu, Haolin and Zhao, Zibo and Lin, Qingxiang and Shi, Huiwen and Yang, Xianghui and Yang, Mingxin and Yang, Shuhui and Feng, Yifei and others},
  journal={arXiv preprint arXiv:2506.16504},
  year={2025}
}

@article{zhong2000object,
  title={Object tracking using deformable templates},
  author={Zhong, Yu and Jain, Anil K and Dubuisson-Jolly, M-P},
  journal={IEEE transactions on pattern analysis and machine intelligence},
  volume={22},
  number={5},
  pages={544--549},
  year={2000},
  publisher={IEEE}
}

@inproceedings{rhodin2015versatile,
  title={A versatile scene model with differentiable visibility applied to generative pose estimation},
  author={Rhodin, Helge and Robertini, Nadia and Richardt, Christian and Seidel, Hans-Peter and Theobalt, Christian},
  booktitle={Proceedings of the IEEE international conference on computer vision},
  pages={765--773},
  year={2015}
}

@inproceedings{rhodin2016general,
  title={General automatic human shape and motion capture using volumetric contour cues},
  author={Rhodin, Helge and Robertini, Nadia and Casas, Dan and Richardt, Christian and Seidel, Hans-Peter and Theobalt, Christian},
  booktitle={European conference on computer vision},
  pages={509--526},
  year={2016},
  organization={Springer}
}

@article{plankers_articulated_2003,
  author = {Plankers, R. and Fua, P.},
  journal = {PAMI},
   number = 9,
  pages = {1182--1187},
  title = {Articulated soft objects for multiview shape and motion capture},
   volume = 25,
  year = 2003
}

@article{Moeslund2006,
 author = {Moeslund, Thomas B. and Hilton, Adrian and Kr\"{u}ger, Volker},
 title = {A Survey of Advances in Vision-based Human Motion Capture and Analysis},
 journal = {CVIU},
 volume = {104},
 number = {2},
 year = {2006},
 issn = {1077-3142},
 pages = {90--126},
 doi = {10.1016/j.cviu.2006.08.002},
 acmid = {1225846},
}

@inproceedings{Bretzner2002,
author = {Bretzner, L. and Laptev, I. and Lindeberg, T.},
doi = {10.1109/AFGR.2002.1004190},
isbn = {0-7695-1602-5},
booktitle = {Automatic Face and Gesture Recognition},
pages = {423--428},
title = {Hand gesture recognition using multi-scale colour features, hierarchical models and particle filtering},
year = {2002}
}

@inproceedings{Sridhar2015,
 author = {Sridhar, Srinath and Mueller, Franziska and Oulasvirta, Antti and Theobalt, Christian},
 title = {Fast and Robust Hand Tracking Using Detection-Guided Optimization},
 booktitle = {CVPR},
 url = {http://handtracker.mpi-inf.mpg.de/projects/FastHandTracker/},
 year = {2015}
}

@inproceedings{Elhayek2015,
	author = {A. Elhayek and E. Aguiar and A. Jain and J. Tompson and L. Pishchulin and M. Andriluka and C. Bregler and B. Schiele and C. Theobalt},
	title = {Efficient {ConvNet}-based Marker-less Motion Capture in General Scenes with a Low Number of Cameras},
	booktitle = {CVPR},
	year = {2015}
}

@inproceedings{Ren2014,
author = {Ren, Carl Yuheng and Prisacariu, Victor and Kaehler, Olaf and Reid, Ian and Murray, David},
doi = {10.1109/3DV.2014.39},
isbn = {978-1-4799-7000-1},
booktitle = {3DV},
pages = {47--54},
title = {{3D} Tracking of Multiple Objects with Identical Appearance Using {RGB-D} Input},
url = {http://ieeexplore.ieee.org/lpdocs/epic03/wrapper.htm?arnumber=7035808},
year = {2014}
}

@article{rhodin2016egocap,
  title={Egocap: egocentric marker-less motion capture with two fisheye cameras},
  author={Rhodin, Helge and Richardt, Christian and Casas, Dan and Insafutdinov, Eldar and Shafiei, Mohammad and Seidel, Hans-Peter and Schiele, Bernt and Theobalt, Christian},
  journal={ACM Transactions on Graphics (TOG)},
  volume={35},
  number={6},
  pages={1--11},
  year={2016},
  publisher={ACM New York, NY, USA}
}

@inproceedings{rajivc2025multi,
  title={Multi-view 3d point tracking},
  author={Raji{\v{c}}, Frano and Xu, Haofei and Mihajlovic, Marko and Li, Siyuan and Demir, Irem and G{\"u}ndo{\u{g}}du, Emircan and Ke, Lei and Prokudin, Sergey and Pollefeys, Marc and Tang, Siyu},
  booktitle={Proceedings of the IEEE/CVF International Conference on Computer Vision},
  pages={59--68},
  year={2025}
}

@article{cong2025dytact,
  title={DyTact: Capturing Dynamic Contacts in Hand-Object Manipulation},
  author={Cong, Xiaoyan and Xing, Angela and Pokhariya, Chandradeep and Fu, Rao and Sridhar, Srinath},
  journal={arXiv preprint arXiv:2506.03103},
  year={2025}
}

@inproceedings{liugeneralizable,
  title={Generalizable Hand-Object Modeling from Monocular RGB Images via 3D Gaussians},
  author={Liu, Xingyu and Ren, Pengfei and Qi, Qi and Sun, Haifeng and Zhuang, Zirui and Wang, Jing and Liao, Jianxin and Wang, Jingyu},
  booktitle={The Thirty-ninth Annual Conference on Neural Information Processing Systems},
year={2025}
}

@inproceedings{sridhar2014real,
  title={Real-time hand tracking using a sum of anisotropic gaussians model},
  author={Sridhar, Srinath and Rhodin, Helge and Seidel, Hans-Peter and Oulasvirta, Antti and Theobalt, Christian},
  booktitle={2014 2nd International Conference on 3D Vision},
  volume={1},
  pages={319--326},
  year={2014},
  organization={IEEE}
}

@inproceedings{banerjee2025hot3d,
  title={Hot3d: Hand and object tracking in 3d from egocentric multi-view videos},
  author={Banerjee, Prithviraj and Shkodrani, Sindi and Moulon, Pierre and Hampali, Shreyas and Han, Shangchen and Zhang, Fan and Zhang, Linguang and Fountain, Jade and Miller, Edward and Basol, Selen and others},
  booktitle={Proceedings of the IEEE/CVF Conference on Computer Vision and Pattern Recognition},
  pages={7061--7071},
  year={2025}
}

@misc{gemini,
  author       = {{Google}},
  title        = {{Gemini 3 Flash Preview}},
  year         = {2025},
  howpublished = {{\url{https://ai.google.dev/gemini-api/docs/models/gemini-3-flash-preview}}},
  note         = {{Model identifier: \texttt{gemini-3-flash-preview}}}
}

@article{simtoolreal,
  title={SimToolReal: An Object-Centric Policy for Zero-Shot Dexterous Tool Manipulation},
  author={Kedia, Kushal and Lum, Tyler Ga Wei and Bohg, Jeannette and Liu, C Karen},
  journal={arXiv preprint arXiv:2602.16863},
  year={2026}
}

@article{shi2026agile,
  title={AGILE: Hand-Object Interaction Reconstruction from Video via Agentic Generation},
  author={Shi, Jin-Chuan and Ye, Binhong and Liu, Tao and Liu, Xiaoyang and Xu, Yangjinhui and He, Junzhe and Li, Zeju and Chen, Hao and Shen, Chunhua},
  journal={arXiv preprint arXiv:2602.04672},
  year={2026}
}

@inproceedings{park2022handoccnet,
  title={Handoccnet: Occlusion-robust 3d hand mesh estimation network},
  author={Park, JoonKyu and Oh, Yeonguk and Moon, Gyeongsik and Choi, Hongsuk and Lee, Kyoung Mu},
  booktitle={Proceedings of the IEEE/CVF conference on computer vision and pattern recognition},
  pages={1496--1505},
  year={2022}
}

@inproceedings{xu2023h2onet,
  title={H2onet: Hand-occlusion-and-orientation-aware network for real-time 3d hand mesh reconstruction},
  author={Xu, Hao and Wang, Tianyu and Tang, Xiao and Fu, Chi-Wing},
  booktitle={Proceedings of the IEEE/CVF conference on computer vision and pattern recognition},
  pages={17048--17058},
  year={2023}
}

@inproceedings{lucas1981iterative,
  title={An iterative image registration technique with an application to stereo vision},
  author={Lucas, Bruce D and Kanade, Takeo},
  booktitle={IJCAI'81: 7th international joint conference on Artificial intelligence},
  volume={2},
  pages={674--679},
  year={1981}
}

@inproceedings{darcet2024vision,
  title={Vision transformers need registers},
  author={Darcet, Timoth{\'e}e and Oquab, Maxime and Mairal, Julien and Bojanowski, Piotr},
  booktitle={International Conference on Learning Representations},
  volume={2024},
  pages={2632--2652},
  year={2024}
}

@article{dinov2,
  title={Dinov2: Learning robust visual features without supervision},
  author={Oquab, Maxime and Darcet, Timoth{\'e}e and Moutakanni, Th{\'e}o and Vo, Huy and Szafraniec, Marc and Khalidov, Vasil and Fernandez, Pierre and Haziza, Daniel and Massa, Francisco and El-Nouby, Alaaeldin and others},
  journal={arXiv preprint arXiv:2304.07193},
  year={2023}
}

@inproceedings{hu2022physical,
  title={Physical interaction: Reconstructing hand-object interactions with physics},
  author={Hu, Haoyu and Yi, Xinyu and Zhang, Hao and Yong, Jun-Hai and Xu, Feng},
  booktitle={SIGGRAPH Asia 2022 conference papers},
  pages={1--9},
  year={2022}
}

@inproceedings{hu2024hand,
  title={Hand-object interaction controller (hoic): Deep reinforcement learning for reconstructing interactions with physics},
  author={Hu, Haoyu and Yi, Xinyu and Cao, Zhe and Yong, Jun-Hai and Xu, Feng},
  booktitle={Acm siggraph 2024 conference papers},
  pages={1--10},
  year={2024}
}

@article{chen2026forehoi,
  title={ForeHOI: Feed-forward 3D Object Reconstruction from Daily Hand-Object Interaction Videos},
  author={Chen, Yuantao and Chang, Jiahao and Ye, Chongjie and Zhang, Chaoran and Fang, Zhaojie and Li, Chenghong and Han, Xiaoguang},
  journal={arXiv preprint arXiv:2602.06226},
  year={2026}
}

@inproceedings{zwicker2001ewa,
  title={Ewa volume splatting},
  author={Zwicker, Matthias and Pfister, Hanspeter and Van Baar, Jeroen and Gross, Markus},
  booktitle={Proceedings Visualization, 2001. VIS'01.},
  pages={29--538},
  year={2001},
  organization={IEEE}
}

@ARTICLE{zwicker2002ewa,
  author={Zwicker, M. and Pfister, H. and van Baar, J. and Gross, M.},
  journal={IEEE Transactions on Visualization and Computer Graphics}, 
  title={EWA splatting}, 
  year={2002},
  volume={8},
  number={3},
  pages={223-238},
  doi={10.1109/TVCG.2002.1021576}
}

@article{haco,    
title = {Learning Dense Hand Contact Estimation from Imbalanced Data},
author = {Jung, Daniel Sungho and Lee, Kyoung Mu},
journal = {Advances in Neural Information Processing Systems},  
year = {2025}  
}

@inproceedings{chao2021dexycb,
  title={Dexycb: A benchmark for capturing hand grasping of objects},
  author={Chao, Yu-Wei and Yang, Wei and Xiang, Yu and Molchanov, Pavlo and Handa, Ankur and Tremblay, Jonathan and Narang, Yashraj S and Van Wyk, Karl and Iqbal, Umar and Birchfield, Stan and others},
  booktitle={2021 IEEE/CVF Conference on Computer Vision and Pattern Recognition (CVPR)},
  pages={9040--9049},
  year={2021},
  organization={IEEE}
}

\clearpage
\section{Supplementary Material}

For consistency across datasets and comparison methods, we resample all input videos to \textbf{10 fps} prior to preprocessing.

\subsection{Additional Evaluation}
Following ARCTIC~\cite{fan2023arctic}, we also report interaction metrics in \cref{tab:interaction_metrics} such as Contact Deviation (CDev) that measures hand-object contact accuracy, Acceleration Error (ACC) to measure motion smoothness, Mean Relative-Root Position Error (MRRPE) to measure the root translation between hand and object, and Motion Deviation (MDev) to check when a hand moves an object, if the vertices of the hand and the object in stable contact move together.
We don't report Average Articulation Error (AAE) since it is for object articulation, and articulated objects are out of our scope.
For detailed explanations of these metrics, please check ARCTIC~\cite{fan2023arctic}.
Please note that, for object acceleration error, we use a topology-agnostic variant rather than the original ARCTIC definition.
Since different methods often output different object meshes, predicted and ground-truth meshes generally do not share vertex correspondence or indexing.
Therefore, we measure object acceleration error by comparing the accelerations of object root trajectories over time.

Table~\ref{tab:interaction_metrics} shows that GraG consistently improves interaction consistency, with the strongest gains on HOT3D.
On HO3Dv3, our method achieves the best temporal interaction quality (best MDev and ACC$_h$), while remaining close to the strongest contact/pose baseline HOLD.
This shows that our predictions are not only accurate in static contact proximity, but also more coherent over time during manipulation.

On HOT3D, the margin is larger: we substantially outperform prior methods in CDev, MDev, and MRRPE$_{h\to o}$.
In particular, CDev drops from the 451--570\,mm range of HOLD/BIGS to 23.8\,mm, and MRRPE$_{h\to o}$ drops from 606--762\,mm to 155\,mm.
These improvements indicate better hand-object coupling under more challenging conditions (dynamic camera, larger motion, and noisier geometry).
HOT3D setting is particularly difficult for the comparison methods due to them depending on COLMAP/SfM initializations.
ACC$_o$ shows a complementary trend.
MagicHOI obtains lower object acceleration error but with much worse CDev/MRRPE$_{h\to o}$ because its object trajectory is often overly smooth in long sequences. 
When COLMAP/SfM fails, the recovered camera/object motion can become nearly static, producing low acceleration despite incorrect object pose. 
Thus ACC$_o$ alone does not indicate accurate object tracking.
Our method GraG provides a better balance: low ACC$_o$ together with strong contact and relative-pose consistency, which is more desirable for physically plausible manipulation.

\paragraph{Number of Gaussians trade-off.}
Increasing the Gaussian count substantially increases compute while giving little measurable improvement on this ablation set.
Moving from Sparse SoG to the reduced SAM3D Gaussian asset increases runtime by about $4.44\times$ and peak CUDA memory by about $2.3\times$, but changes reconstruction metrics only marginally.
Using the full dense Gaussian asset is much more expensive: it is over two orders of magnitude slower than Sparse SoG, while yielding no meaningful gain in CD, F10, MPJPE, or hand-object reconstruction quality.
This supports our design choice of optimizing a compact sparse isotropic SoG: it preserves the useful geometric signal from SAM3D while avoiding the optimization cost of tracking tens or hundreds of thousands of Gaussians.
  \begin{table}[t]
  \centering
  \caption{
 \textbf{ Trade-off between Gaussian count, compute, memory, and reconstruction quality} on the ablation set.
  Runtime is normalized to hours per 100 frames as in the main paper setting.
  ``Sparse SoG'' is our default setting where we use farthest point sampling to subsample from the reduced SAM3D Gaussian asset and convert to isotropic Gaussians.
  ``Reduced'' denotes SAM3D's lower-count Gaussian decoder output, trained for downstream needs such as decoding fewer Gaussians for faster inference (see SAM3D's Supplementary Material for detailed information); ``Full'' denotes the dense SAM3D Gaussian decoder output. 
  Storage reports the memory occupied by the object Gaussian parameter tensors, while Peak CUDA reports the maximum CUDA memory allocated during optimization.
  Thus, Storage measures representation compactness, whereas Peak CUDA captures practical optimization-time GPU memory.
  }
  \label{tab:gaussian_tradeoff}
  \resizebox{\linewidth}{!}{
  \begin{tabular}{lrrrrrrrr}
  \toprule
  Setting & Number of Gaussians & Runtime (h) & Storage (MB) & Peak CUDA (GB) & CD $\downarrow$ & F10 $\uparrow$ & MPJPE $\downarrow$ & CD$_h$ $\downarrow$ \\
  \midrule
  Sparse SoG          & $\sim 2{,}000$  & 0.27  & 0.11 & 2.4 & 0.45 & 98.8 & 11.3 & 7.65 \\
  Reduced & $\sim 60{,}000$  & 1.20 & 3.20 & 5.5 & 0.44 & 99.1 & 11.3 & 7.50 \\
  Full  & $\sim 300{,}000$ & 34.1 & 16.0 & 6.2 & 0.44 & 99.1 & 11.3 & 7.50 \\
  \bottomrule
  \end{tabular}
  }
  \end{table}

\subsection{Hyperparameters}
\label{sup_sec:hyperparams}
We report the concrete settings used by our method for the components described in \cref{sec:stage3_sog}.

\begin{table}[t]
\caption{
\textbf{Interaction-consistency metrics.}
We report interaction metrics from ARCTIC~\cite{fan2023arctic} on HO3Dv3 and HOT3D, including
Contact Deviation (CDev), Motion Deviation (MDev),
hand/object acceleration error (ACC$_h$, ACC$_o$),
and Mean Relative Root Position Error between the hand and object
(MRRPE$_{h\to o}$).
Lower is better for all metrics.
}
\centering
\small
\resizebox{\columnwidth}{!}{%
\begin{tabular}{llccccc}
\toprule
Dataset & Method
& CDev [mm] $\downarrow$
& MDev [mm] $\downarrow$
& ACC$_h$ [m/s$^2$] $\downarrow$
& ACC$_o$ [m/s$^2$] $\downarrow$
& MRRPE$_{h\to o}$ [mm] $\downarrow$ \\
\midrule

\multirow{4}{*}{HO3Dv3}
& HOLD~\cite{fan2024hold} & \best{15.8}& \second{10.8}& \second{11.3}& 30.2& \best{38.6}\\
& BIGS~\cite{on2025bigs} & 54.2& 23.1& 14.4& 37.8& 79.4\\
& MagicHOI~\cite{wang2025magichoi} & 533& 21.5& 15.4& \best{10.3}& 747\\
& \textbf{Ours} & \second{17.0}& \best{5.37}& \best{6.17}& \second{12.5}& \second{45.4}\\

\midrule

\multirow{4}{*}{HOT3D}
& HOLD~\cite{fan2024hold} & \second{451}& \second{65.3}& \second{15.8}& 117& \second{606}\\
& BIGS~\cite{on2025bigs} & 570& 85.6& 24.1& 136& 762\\
& MagicHOI~\cite{wang2025magichoi} & 463& 66.7& 20.3& \best{18.3}& 728\\
& \textbf{Ours} & \best{23.8}& \best{15.7}& \best{10.5}& \second{27.4}& \best{155}\\

\bottomrule
\end{tabular}%
}
\label{tab:interaction_metrics}
\end{table}

\paragraph{Windowed optimization.}
Stage~3 uses windows of $|\mathcal{W}|=8$ frames with stride $7$ (one frame overlap).  Each window is optimized for $100$ AdamW steps.  We refine object pose and scale jointly with small hand-pose and hand-translation updates, while keeping the solution close to the SAM3D and Dyn-HaMR initialization through the priors in \cref{eq:final_obj}.

\paragraph{Optimizer settings.}
We use AdamW with $\beta=(0.9,0.95)$ and weight decay $0$, with per-variable learning rates: object translation $2.5{\times}10^{-3}$, object rotation $8{\times}10^{-4}$, object scale $10^{-3}$, hand articulated pose $4{\times}10^{-4}$, hand translation $9{\times}10^{-4}$, hand global orientation $5{\times}10^{-4}$, and hand scale $2{\times}10^{-4}$.
The object and hand shapes are fixed during Stage~3.
We do not use gradient clipping; stability instead comes from value clamping after each step: the single global object scale is clamped to $[0.75,\,2.0]\times$ its initialization and the hand scale to $[0.98,\,1.04]\times$ its initialization, so large contact/depth corrections are absorbed by translation rather than scale.
The object scale and the hand scale are single scalars shared over the whole sequence (not a per-window or per-frame parameter).

\paragraph{Number of keyframes.} We use $K=4$ keyframes for object reconstruction.

\paragraph{Image SoG construction.}
The image SoG is built inside the object mask bounding box using a quadtree with maximum depth $6$, color variance threshold $0.04$, minimum cell size $8$ pixels, bounding-box padding $2$ pixels, and minimum valid-mask ratio $10^{-6}$.  We do not split uniform mask interiors.  The isotropic image standard deviation is the geometric mean of the valid-pixel half extents in each leaf.  RGB values are normalized to $[0,1]$.

\paragraph{Object SoG projection.}
We sample $2000$ object Gaussians by farthest-point sampling from the decoded SAM3D Gaussian asset.  The object SoG uses isotropic 3D radii with factor $\gamma=1.75$ as described in the method details below.  For SoG matching, each projected object primitive considers the $64$ nearest image SoG components. 
The color-kernel bandwidth is $\sigma_c=0.15$.

\paragraph{Loss weights.}
The compact Stage~3 objective in \cref{eq:final_obj} uses the following weights:
\[
\begin{aligned}
\lambda_{\text{sog}} &= 0.05, &
\lambda_{\text{j2d}} &= 0.5, &
\lambda_{\text{depth}} &= 1000, \\
\lambda_{\text{sil}} &= 100, &
\lambda_{\text{contact}} &= 1000, &
\lambda_{\text{smooth}} &= 100 .
\end{aligned}
\]

\paragraph{Depth and contact activation.}
The pointmap depth is used as a robust hand/object depth anchor.  In contact frames, the object depth is additionally tied to the hand depth as in \cref{eq:depth_prior}.  The contact indicator $c^{ho}_t$ is set from the VLM contact flag when available, and otherwise falls back to the mask-proximity test of \cref{sup_sec:contact_validation}; it gates the contact and depth-coupling terms. 
Please note that we used VLM contact flags throughout the experiments in the main paper. Mask-proximity flags are only used for contact-signal validation in \cref{tab:contact_validation}.

\begin{table}[t]
\centering
\caption{{\textbf{Contact-signal validation} against mesh-proximity ground truth. Balanced accuracy (Bal.\ Acc) averages the in-contact and no-contact class accuracies; the VLM flag is the stronger signal on both datasets, most clearly on the rare no-contact frames.}}
\label{tab:contact_validation}
\setlength{\tabcolsep}{5pt}
\renewcommand{\arraystretch}{1.1}
\resizebox{\columnwidth}{!}{%
\begin{tabular}{lcccc}
\toprule
& \multicolumn{2}{c}{HO3Dv3} & \multicolumn{2}{c}{HOT3D} \\
\cmidrule(lr){2-3}\cmidrule(lr){4-5}
Contact signal & F1 $\uparrow$ & Bal.\ Acc $\uparrow$ & F1 $\uparrow$ & Bal.\ Acc $\uparrow$ \\
\midrule
Gemini~3 (VLM, ours)      & \textbf{0.99} & \textbf{0.80} & 0.97 & \textbf{0.66} \\
Mask-proximity (fallback) & 0.96 & 0.70 & \textbf{0.98} & 0.59 \\
\bottomrule
\end{tabular}%
}
\vspace{-6pt}
\end{table}

\paragraph{Contact flag from the video VLM}
We compute the contact flag once per sequence with \texttt{gemini-3-flash-preview}~\cite{gemini}.
Instead of labeling frames one at a time, we pass the whole video to the model and ask when the hand is grasping the object, as a list of time intervals, and convert these intervals to per-frame flags $c^{ho}_t$ using the sequence frame rate.
We ask for intervals rather than per-frame labels because the start and end of a grasp are placed more consistently when the model reasons over the full sequence, which also removes the frame-to-frame flicker we saw with per-frame queries under occlusion.
We use the following prompt, where \texttt{\{subject\}} is the interacting hand:

{\footnotesize\begin{verbatim}
Analyze this video frame-by-frame and decide EXACTLY when {subject} 
transitions from NOT holding (0) to HOLDING (1).

CRITICAL RULES:
1. Do NOT assume grasping at 0.0s unless the first visible frame clearly 
   shows the fingers already around the object.
2. Reaching toward the object is NOT grasping.
3. Grasping begins ONLY at the FIRST frame where the fingers visibly enclose 
   the object.
4. If uncertain, DELAY the start until enclosure is visually unambiguous.

Definitions:
- REACHING (0): {subject} moving toward or touching the object without 
  enclosure.
- GRASPING (1): fingers of {subject} clearly wrapped around the object, 
  or the object is clearly lifted while enclosed.

Step 1: Describe the timeline of states in order.
Step 2: Give the FIRST timestamp where grasping begins.
Step 3: Give when grasping ends (if it ends).

Return ONLY a JSON object in a ```json block:
{ "timeline_reasoning": string,
  "contact_intervals":
     [ {"start_sec": num, "end_sec": num} ],
  "no_contact_before_sec": num }

Also provided: duration_sec, sampled_num_frames, sampled_fps.
- Use exact timestamps; do not round aggressively.
- Do not assume holding from 0.0s unless visible.
- If grasping never occurs, return an empty list.
\end{verbatim}}

\subsection{Contact Signal Validation}
\label{sup_sec:contact_validation}
Our contact gate $c^{ho}_t$ (\cref{eq:final_obj}) is obtained from a video VLM (Gemini~3~\cite{gemini}), which we prompt to identify per-sequence contact time intervals, which we then convert to per-frame flags.
We validate this signal against ground-truth contact and against an image-space alternative for reproducibility.

Following ARCTIC~\cite{fan2023arctic}, we label a frame as in-contact when the minimum hand-to-object vertex distance from the ground-truth meshes is below or equal to $3\,$mm, using each dataset's registered hand and object meshes.

As a VLM-free fallback for reproducibility, we also derive contact directly from segmentation masks: at each frame we dilate the binary hand and object masks with a $3{\times}3$ max-pooling kernel (with stride 1) and flag contact when the dilated masks overlap in image space.

Since the clips are contact-heavy ($\sim$97\% of frames in contact), raw accuracy is misleading, so we report F1 and balanced accuracy (the mean of the in-contact and no-contact class accuracies), with the mask dilation chosen to maximize the baseline's balanced accuracy.

The VLM flag is the more reliable signal on both datasets (\cref{tab:contact_validation}).
The gap is clearest on the rare no-contact frames, which balanced accuracy captures: $0.80$ vs.\ $0.70$ on HO3Dv3 and $0.66$ vs.\ $0.59$ on HOT3D.
Mask-proximity reaches a slightly higher F1 on HOT3D only because it labels almost every frame as contact.
It struggles on the no-contact frames because it cannot see depth: as the hand approaches or releases the object, the two masks still overlap in the image while the surfaces are apart in depth, so it reports contact when there is none and would wrongly switch on our depth-coupling loss.
The effect carries into reconstruction: swapping the VLM flag for the tuned mask-proximity flag raises $CD_h$ from $5.02\pm1.47$ to $5.53\pm2.32\,$cm on HO3Dv3 and from $9.37\pm4.68$ to $10.60\pm6.50\,$cm on HOT3D.
A tuned mask-proximity flag thus recovers most of the accuracy, but the VLM flag is better on the frames that matter and needs no tuning.
The VLM flag can still hallucinate contact, but only on $1.2\%$ of HO3Dv3 and $1.8\%$ of HOT3D frames.
Exploring open-source video VLMs or dense hand-contact predictors (e.g., HACO~\cite{haco}) as reproducible contact cues is an interesting future direction.

\subsection{Method Details}
\paragraph{Keyframe Selection Details}
\label{supp:keyframe_selection_details}
To reconstruct a high-quality 3D object, we must select views that are both informative (reliable and stable) and diverse (covering different object angles).
Following the saddle-balanced selection strategy of Depth Anything 3 (DA3)~\cite{depthanything3}, we extract a global descriptor $f_t \in \mathbb{R}^d$ for each frame from the \texttt{[CLS]} token of the DA3 encoder, which uses a DINOv2 backbone~\cite{dinov2}. 
We compute the cosine similarity matrix $S_{t,s}$ between all pairs of frame descriptors and characterize each frame using three scalar statistics: the mean similarity to the remaining frames,
$\bar{S}_t = \frac{1}{T-1}\sum_{s\neq t} S_{t,s}$, the feature magnitude $n_t=\|f_t\|_2$, and the feature variance $\nu_t=\mathrm{Var}(f_t)$ across descriptor dimensions.

Each statistic is independently min--max normalized over the sequence.
We then define a balance cost that favors frames near the center of the normalized feature distribution:
$\mathcal{C}_{\mathrm{bal}}(t)
=
\left|\bar{S}_t - 0.5\right|
+
\left|n_t - 0.5\right|
+
\left|\nu_t - 0.5\right|.$
This midpoint discourages frames with extreme class tokens. 
In particular, unusually large feature norms or variances may reflect unstable activations in ViT features, a phenomenon also discussed in the context of register tokens~\cite{darcet2024vision}, while very small values tend to correspond to frames with weak or ambiguous visual signal. 
The balance cost therefore provides a simple criterion for selecting frames depending on their class tokens.

To avoid selecting redundant views, we combine this cost term with a diversity penalty:
\begin{equation}
    \mathcal{V}^*
    =
    \arg\min_{\mathcal{V}\subset\mathbf{I},\,|\mathcal{V}|=K}
    \sum_{t\in\mathcal{V}}
    \left(
    \mathcal{C}_{\mathrm{bal}}(t)
    +
    \lambda_{\mathrm{div}}
    \max_{s\in\mathcal{V}\setminus\{t\}} S_{t,s}
    \right),
\end{equation}
where $\lambda_{\mathrm{div}}$ controls the trade-off between stability and diversity. We optimize this objective greedily: the first keyframe is chosen as the frame with the lowest cost, and each subsequent keyframe is selected by minimizing
\begin{equation}
\mathcal{C}_{\mathrm{bal}}(t)
+
\lambda_{\mathrm{div}}
\max_{s\in\mathcal{V}} S_{t,s},
\qquad t\notin\mathcal{V}.
\end{equation}
This procedure selects frames whose descriptors are representative of the sequence, while avoiding views that are too similar to the keyframes already selected.
Example keyframes can be seen in \cref{fig:stage1}.

\paragraph{Flow Matching Guidance Details}
\label{supp:sam3d_details}
In Stage 2, we modify the internal generative process of SAM3D~\cite{sam3d} to achieve temporal stability. 
\begin{itemize}
    \item \textbf{Latent Space and Projections:} SAM3D represents object shape ($Z^S$) and layout ($Z^P$) as sets of discrete tokens. Prior to being processed by the SAM3D transformer, these tokens are mapped into a shared high-dimensional feature space via modality-specific \textit{input projection} layers (MLPs). Let $X^P_t \in \mathbb{R}^{D}$ denote the layout latent state for frame $t$, where the dimension $D$ is typically 1024. 
    \item \textbf{Guidance in the Velocity Field:} SAM3D performs 3D reconstruction using \textit{Conditional Flow Matching}. In this paradigm, the transformer does not predict the final pose directly; rather, it predicts a velocity field $v^P_\theta$ that defines the direction the current latent state $X^P_t$ should move within the feature space to reach the target data distribution.
    \item \textbf{Guidance in the Velocity Field:} Our temporal guidance mechanism is applied directly within this 1024-dimensional space. Because velocity $v^P_\theta = \frac{dX^P_t}{dt}$ represents a displacement vector in the feature space, it is dimensionally equivalent to the latent state $X^P_t$. We compute the guided velocity $\tilde{v}^P_\theta$ as:
\begin{equation}
\tilde{v}^{P}_\theta \;=\; v^{P}_\theta \;-\; \lambda_{\text{temp}}\,(X^{P}_t - X^{P}_{t-1}),
\end{equation}
where $(X^{P}_t - X^{P}_{t-1})$ acts as a \textit{temporal spring} that discourages abrupt changes in orientation or translation between consecutive frames.
\item \textbf{Output Projection and Token Update:} Following the guidance step, the guided velocity $\tilde{v}^P_\theta$ is passed through a modality-specific \textit{output projection} MLP. This layer maps the 1024D velocity back into the original token space, yielding a token-space update $\Delta Z$. This update is applied to the tokens of the previous frame:
\begin{equation}
Z^P_t \;=\; Z^P_{t-1} \;+\; \Delta Z.
\end{equation}
The updated tokens $Z^P_t$ are then processed by the Layout Decoder to produce the physical rotation, translation, and scale\cite{sam3d}. By performing guidance at the latent velocity level, the model's self-attention mechanisms can resolve geometric ambiguities while adhering to a strong prior on temporal continuity.
\end{itemize}
At the end of Stage~2, there might still be some occasional orientation flips in the outputs.
We perform a simple quaternion consistency check and reverse the out-of-order flips to handle them.

\paragraph{Image SoG}
\label{sup_sec:image_sog_details}
The image SoG used in \cref{sec:image_sog} is built over the tight object-mask bounding box.
Each quadtree leaf $\Omega_i$ stores only the pixels belonging to the object mask, $\mathcal{P}_i=\{u\in\Omega_i\mid M^o_t(u)=1\}$.
Its center, color, and weight are computed as
\[
\mu_i=\frac{1}{|\mathcal{P}_i|}\sum_{u\in\mathcal{P}_i}u,
\qquad
c_i=\frac{1}{|\mathcal{P}_i|}\sum_{u\in\mathcal{P}_i}I_t(u),
\qquad
w_i=|\mathcal{P}_i|.
\]
For an isotropic footprint, we compute the horizontal and vertical extents of valid pixels in the leaf, $r^x_i$ and $r^y_i$, and set $\sigma_i=\sqrt{r^x_i r^y_i}$.
The optimized image representation is therefore the compact set $\mathcal{K}_{I_t}=\{(\mu_i,\sigma_i,c_i,w_i)\}_{i=1}^{N_t}$.
For visualization only, we rasterize these components back to the image plane with Gaussian-weighted color compositing.

\paragraph{Object SoG}
\label{sup_sec:object_sog_details}
The object SoG used in \cref{sec:gauss_to_sog} is obtained by sparsifying the decoded SAM3D Gaussian asset with farthest-point sampling and converting each retained anisotropic Gaussian to an isotropic tracking primitive.
Let $a_j\in\mathbb{R}^3$ denote the decoded Gaussian scale.
We use the 3D radius
\[
\rho_j = \gamma\,\frac{a^x_j+a^y_j+a^z_j}{3},
\]
with $\gamma=1.75$, yielding $\mathcal{G}_o=\{(x_j,\rho_j,c_j)\}_{j=1}^{N_o}$ and $N_o\le 2000$.
Under the current object pose and camera, each primitive projects to
\[
\mu^{2D}_{j,t}=\Pi_t(s^o R^o_t x_j + \tau^o_t + \Delta\tau^o_t),
\qquad
\sigma^{2D}_{j,t}=
\frac{\bar f_t\,s^o\rho_j}{z_{j,t}},
\]
where $\bar f_t=(f^x_t+f^y_t)/2$ and $z_{j,t}$ is camera depth.

\paragraph{Color kernel for SoG}
\label{color_kernel}
For two color vectors $c_i,c_j\in\mathbb{R}^3$, with RGB normalized to $[0,1]$, we define
\begin{equation}
d(c_i,c_j)
=
\exp\!\left(-\frac{\lVert c_i-c_j\rVert_2^2}{\sigma_c^2}\right),
\label{eq:color_kernel}
\end{equation}
where $\sigma_c$ controls how strongly color differences are penalized.
We set $\sigma_c=0.15$ in all experiments.

\paragraph{Stage 3 prior losses.}
\label{sup_sec:prior_losses}
This subsection gives the concrete auxiliary losses used with the SoG alignment term in \cref{eq:final_obj}.
Since the initialization from Dyn-HaMR already provides strong hand tracking, we do not use SoG alignment for the hand. 
Instead, we regularize the hand trajectory with a lightweight 2D joint reprojection prior.
Using the MANO joint regressor $J$, we obtain the 3D location of joint $\ell$ via $J_{t,\ell}(\boldsymbol{\Theta}_t,{R}^h_t,{\tau}^h_t, s^h)\in\mathbb{R}^3$. Given detected 2D joints $\hat{u}_{t,\ell}$, the reprojection prior is
\begin{equation}
\mathcal{L}_{\text{j2d}} =
\sum_{t\in \mathcal{W}}
\sum_{\ell}
\left\|
\Pi\!\big(J_{t,\ell}(\boldsymbol{\Theta}_t,R^h_t,\tau^h_t,s^h)\big)
-
\hat{u}_{t,\ell}
\right\|_2^2.
\label{eq:j2d}
\end{equation}

For the depth loss, although pointmaps provide useful geometric cues, we use them only as aggregate anchors.  In hand-object videos, object-side pointmaps are often noisy near occlusions or when the visible object region is small.  A dense pointmap loss can then bias object pose or scale.  We therefore use the compact depth term in \cref{eq:depth_prior}: pointmap statistics anchor metric depth, while contact frames tie the object depth to the hand depth.

For the contact loss, let $\mathcal{C}_t$ be the selected MANO contact-zone vertices in frame $t$, and let $\mathcal{V}(H_t)$ and $\mathcal{V}(O_t)$ denote the vertices of the posed hand and object surfaces.
The contact attraction term is
\[
\mathcal{L}_{\text{near}}
=
\sum_{t\in\mathcal{W}} c_t^{ho}
\frac{1}{|\mathcal{C}_t|}
\sum_{v\in\mathcal{C}_t}
\min_{y\in O_t}\|v-y\|_2^2 .
\]
To penalize interpenetration, let $\phi_{O_t}(x)$ and $\phi_{H_t}(x)$ be signed distances to the object and hand surfaces, positive outside and negative inside.
We use
\[
\begin{split}
\mathcal{L}_{\text{pen}}
=
\sum_{t\in\mathcal{W}} c_t^{ho}
\bigg(
&\frac{1}{|\mathcal{V}(H_t)|}
\sum_{v\in\mathcal{V}(H_t)}
\max(0,-\phi_{O_t}(v))^2\\
&+
\frac{1}{|\mathcal{V}(O_t)|}
\sum_{y\in\mathcal{V}(O_t)}
\max(0,-\phi_{H_t}(y))^2
\bigg).
\end{split}
\]

The full contact loss is $\mathcal{L}_{\text{contact}}=\mathcal{L}_{\text{near}}+\lambda_{\text{pen}}\mathcal{L}_{\text{pen}}$ where $\lambda_{\text{pen}}=5$.
This combination pulls the active fingertip/contact region toward the object surface while discouraging either surface from passing through the other.

For temporal stability, we penalize hand/object translation acceleration and object angular acceleration:
\[
\mathcal{L}_{\text{smooth}}
=
\sum_{t\in\mathcal{W}}
\left(
\|\Delta^2 \tau^h_t\|_2^2
+
\|\Delta^2 \tau^o_t\|_2^2
+
\lambda_R\|\Delta^2 \omega^o_t\|_2^2
\right),
\]
where $\omega^o_t$ denotes the rotational velocity induced by consecutive object quaternions.

\subsection{Isotropic vs. Anisotropic Object Gaussian Projection}
\label{supp:aniso}
In this section, we detail the projection approximation and the anisotropic projection ablation with a first-order perspective Jacobian.
\citet{zwicker2001ewa}'s volume derivation maps camera space to three-dimensional ray space with a $3{\times}3$ Jacobian and obtains the 2D footprint covariance by dropping the depth row and column. 
Their surface derivation equivalently uses the direct $2{\times}3$ Jacobian of the camera-to-image map (Eq.~34 in~\cite{zwicker2002ewa}). 
Our implementation uses this direct image-space form.
Let object Gaussian $j$ have camera-space mean $\mu_j^c=(x,y,z)^T$ and isotropic world-space covariance $\sigma_j^2\mathbf{I}_3$. The pixel projection is
\begin{equation}
\mu_j
=
\boldsymbol{\pi}(\mu_j^c)
=
\begin{bmatrix}
f_x x/z+c_x \\ f_y y/z+c_y
\end{bmatrix}.
\end{equation}
Its image-space Jacobian is
\begin{equation}
\mathbf{A}
=
\frac{\partial\boldsymbol{\pi}}{\partial\mu_j^c}
=
\begin{bmatrix}
f_x/z & 0 & -f_x x/z^2 \\
0 & f_y/z & -f_y y/z^2
\end{bmatrix}
\in\mathbb{R}^{2\times3}.
\end{equation}
Note that $\mathbf{A}$ is the derivative of the 2D pixel projection, not the $3{\times}3$ ray-space Jacobian of the volume formulation. 
Let $\mathbf{R}_{wc}$ denote the rotational part of the world-to-camera transform. 
The covariance implemented by the anisotropic branch is
\begin{equation}
\begin{split}
\Sigma_j
&=
(\mathbf{A}\mathbf{R}_{wc})(\sigma_j^2\mathbf{I}_3)
(\mathbf{A}\mathbf{R}_{wc})^T \\
&=
\sigma_j^2\mathbf{A}\mathbf{A}^T,
\end{split}
\label{eq:aniso_cov_projection}
\end{equation}
where the second equality follows because $\mathbf{R}_{wc}$ is orthonormal. 
This covariance can be written as
\begin{equation}
\Sigma_j
=
\sigma_j^2
\begin{bmatrix}
\dfrac{f_x^2}{z^2}\!\left(1+\dfrac{x^2}{z^2}\right)
& \dfrac{f_xf_yxy}{z^4} \\[7pt]
\dfrac{f_xf_yxy}{z^4}
& \dfrac{f_y^2}{z^2}\!\left(1+\dfrac{y^2}{z^2}\right)
\end{bmatrix}.
\end{equation}
Thus, perspective distortion produces an elliptical footprint in general. 
Our isotropic approximation replaces this covariance by a scalar screen-space variance.
The anisotropic ablation uses $\Sigma_j$ for projected model Gaussian $j$, while image Gaussian $i$ remains isotropic with $\Sigma_i=\sigma_i^2\mathbf{I}_2$. 
Following the anisotropic Gaussian overlap of \citet{sridhar2014real}, the continuous spatial overlap of the unnormalized Gaussians is
\begin{equation}
\int_{\Omega}\mathcal{B}_i(u)\mathcal{B}_j(u)\,du
={} 2\pi
\frac{\sqrt{|\Sigma_i|}\sqrt{|\Sigma_j|}}
{\sqrt{|\Sigma_i+\Sigma_j|}}\cdot\mathrm{e}^{-\frac{1}{2}(\mu_i-\mu_j)^T
(\Sigma_i+\Sigma_j)^{-1}
(\mu_i-\mu_j)}.
\label{eq:aniso_sog_overlap}
\end{equation}
Multiplying this spatial overlap by the same color kernel $d(c_i,c_j)$ gives $E_{ij}$ as defined in \cref{eq:eij_def}. 
For isotropic model and image covariances, $\Sigma_j=\sigma_j^2\mathbf{I}_2$ and $\Sigma_i=\sigma_i^2\mathbf{I}_2$, \cref{eq:aniso_sog_overlap} reduces exactly to the scalar expression in \cref{eq:gauss_overlap_iso}.
This scalar form avoids per-pair $2{\times}2$ determinant and inverse evaluations, providing a smoother image alignment signal. 

We repeat our optimization stage (Stage~3) using the described anisotropic projection and general overlap, while all other parameters remain unchanged.
Results are reported in \cref{tab:aniso_proj}.
Under this setting, Jacobian-based anisotropic projection results in a slightly higher $CD_h$ (from $5.02$ to $5.29$ on HO3Dv3, from $9.37$ to $10.43$ on HOT3D) and is ${\sim}11\%$ slower.
\begin{table}[t]
\centering
\caption{{\textbf{Gaussian-projection ablation.} Comparing our scalar closed-form overlap against full 2D anisotropic projection. Modeling Jacobian-based anisotropic projection with matrix overlap does not improve tracking accuracy ($CD_h$) and adds ${\sim}11\%$ runtime.}}
\label{tab:aniso_proj}
\resizebox{\columnwidth}{!}{%
\begin{tabular}{llccc}
\toprule
Projection & Overlap & \shortstack{HO3Dv3\\$CD_h[\mathrm{cm}]\downarrow$} & \shortstack{HOT3D\\$CD_h[\mathrm{cm}]\downarrow$} & \shortstack{Runtime\\$[\mathrm{h}]\downarrow$} \\
\midrule
Isotropic (ours) & Scalar closed-form & $\textbf{5.02}\pm\textbf{1.47}$ & $\textbf{9.37}\pm\textbf{4.68}$ & $\textbf{0.27}$ \\
Anisotropic      & Full matrix ($2{\times}2$) & $5.29\pm2.04$ & $10.43\pm6.39$ & $0.30$ \\
\bottomrule
\end{tabular}%
}
\end{table}

One possible explanation for the lack of improvement in the anisotropic variant's performance is that the spatial arrangement of $2000$ primitives already captures much of the object-level shape needed for pose refinement (including  elongated object structures such as scissors and screwdrivers).
In this setting, modeling a more accurate elliptical footprint provides limited additional information.
The general matrix overlap may also introduce a less smooth alignment landscape than the scalar overlap.
A hybrid formulation of isotropy and anisotropy for boundaries or elongated object regions, and representing image SoG also as anisotropic Gaussians remains as future directions.

\subsection{Evaluated Sequences}
We evaluate our method on sequences from HO3Dv3~\cite{hampali2021ho3dv3} and HOT3D~\cite{banerjee2025hot3d}.
Specifically, we show which sequences and which timestamps used in \cref{tab:ho3d_seq} and \cref{tab:hot3d_seq}.

\begin{table}[h]
\centering
\caption{HO3Dv3 sequences for hand-object reconstruction. We directly use the same sequences from HOLD~\cite{fan2024hold}. All of these sequences are only right-handed.}
\begin{tabular}{ll}
\hline
\textbf{Object} & \textbf{Sequence name} \\
\hline
bleach       & ABF12  \\
bleach       & ABF14  \\
potted meat  & GPMF12 \\
potted meat  & GPMF14 \\
cracker box  & MC1    \\
cracker box  & MC4    \\
power drill  & MDF12  \\
power drill  & MDF14  \\
sugar box    & ShSu10 \\
sugar box    & ShSu12 \\
mustard      & SM2    \\
mustard      & SM4    \\
mug          & SMu1   \\
mug          & SMu40  \\
banana       & BB12   \\
banana       & BB13   \\
scissors     & GSF12  \\
scissors     & GSF13  \\
\hline
\end{tabular}
\label{tab:ho3d_seq}
\end{table}
\begin{table*}[t]
\centering
\small
\caption{HOT3D clips used in our evaluations, reported with sequence identifiers, hand side, temporal boundaries, and object IDs. 
Temporal boundaries are reported as HOT3D timecode timestamps (TS) in nanoseconds.
}
\begin{tabular}{lllll}
\hline
\textbf{Sequence name} & \textbf{Hand side} & \textbf{Start TS} (ns) & \textbf{End TS} (ns) & \textbf{Object ID} \\
\hline
P0001\_23fa0ee8 & right & 55003474858015 & 55015408224119 & 238686662724712 \\
P0001\_8d136980 & right & 54039408208193 & 54044341521984 & 238686662724712 \\
P0001\_b2bcbe28 & left & 54811408205972 & 54814874891157 & 258906041248094 \\
P0001\_f71fc9b1 & right & 66482274870136 & 66485208197200 & 249541253457812 \\
P0010\_8ff3e5c4 & left & 58695141597662 & 58697074869546 & 258906041248094 \\
P0011\_0c2d00ed & right & 55563708197241 & 55572041548328 & 70709727230291 \\
P0011\_0c2d00ed & right & 55660041543187 & 55661741528834 & 204462113746498 \\
P0011\_2255f410 & right & 56349641532049 & 56356641531612 & 249541253457812 \\
P0011\_ccc678c7 & right & 55922341542825 & 55926474871989 & 261746112525368 \\
P0011\_cd22f5e0 & right & 55211841541663 & 55215108194835 & 228358276546933 \\
P0011\_ff7cfc3a & right & 55460874890915 & 55465808213471 & 225397651484143 \\
P0011\_ff7cfc3a & right & 55518174857985 & 55522441524737 & 270231216246839 \\
P0012\_0a21e4c2 & right & 47121508235754 & 47124108213732 & 106957734975303 \\
P0012\_0a21e4c2 & right & 47145441536211 & 47157641593381 & 258906041248094 \\
P0012\_44c5f677 & left & 59321408185000 & 59324774850689 & 194930206998778 \\
P0012\_44c5f677 & right & 59420774854423 & 59423208211436 & 106434519822892 \\
P0014\_84ea2dcc & right & 44789441528790 & 44791608213887 & 98604936546412 \\
P0015\_60573a3b & left & 60819408201515 & 60825874885641 & 79582884925181 \\
\hline
\end{tabular}
\label{tab:hot3d_seq}
\end{table*}

\subsection{Comparison Details}
For HOLD~\cite{fan2024hold}, we employ the authors' publicly available pre-trained CVPR models for HO3D and use the available code to train for HOT3D.
For BIGS~\cite{on2025bigs}, we use the official implementation from the project repository and optimize each stage using the hyperparameters specified in their Supplementary Material (Sec.~A).
To get meshes out of their optimized Gaussians, we employ Poisson Reconstruction with depth $6$ after we remove the outliers within $100$ neighbors with $2.0$ standard deviation.
For MagicHOI~\cite{wang2025magichoi}, we follow the official implementation from the project repository.

\subsection{Detailed Results Discussion}
GraG differs fundamentally from recent hand-object reconstruction methods such as HOLD, BIGS, and MagicHOI, which rely on COLMAP/SfM-based initialization followed by heavy per-scene optimization.
These methods depend on stable multi-view correspondences to estimate camera poses and coarse geometry before performing joint reconstruction.
However, such initialization is often unreliable in hand-object interaction videos due to severe hand occlusions, rapid camera motion, and small object projections, which are particularly common in datasets such as HOT3D.
As a result, errors in the early SfM stage can propagate to later optimization stages, leading to degraded reconstruction quality and lower success rates.

MagicHOI further combines SfM initialization with generative priors and object inpainting to complete unseen regions.
While effective under carefully selected inputs, achieving strong results typically requires manual selection of reconstruction frames and inpainting viewpoints.
When the full video sequence is provided directly without such curation, the optimization becomes significantly more challenging and the reconstruction quality degrades, especially for longer sequences such as those in HO3Dv3 and HOT3D.
In contrast, GraG is designed to operate directly on full video sequences without manual frame selection.
Our method first identifies informative keyframes and reconstructs a canonical object representation using a foundation model, while adapting SAM3D-based object pose prediction to video sequences.
The canonical object is then tracked across the sequence using a compact SoG representation with temporal and contact-aware constraints.
By decoupling canonical shape reconstruction from per-frame motion estimation, GraG avoids repeated large-scale scene optimization and instead performs efficient low-dimensional tracking.

This design yields several advantages.
First, GraG is substantially more robust to occlusion and unstable viewpoints, which explains its stronger performance on challenging datasets such as HOT3D where SfM-based methods often struggle.
Second, reconstructing a canonical object geometry from informative keyframes preserves fine structures, making the approach more reliable for thin or partially observed objects that are difficult to recover through correspondence-based initialization.
Finally, replacing heavy test-time optimization with lightweight tracking over a fixed canonical asset enables significantly faster reconstruction while maintaining higher accuracy on both HO3Dv3 and HOT3D.

\subsection{Why SoG Instead of Rendering or Optical Flow?}
The dense MV-SAM3D Gaussian asset remains important: it defines the canonical geometry and appearance from which the compact object SoG is sampled.
However, GraG does not optimize the full photorealistic asset during tracking.
Stage~3 repeatedly re-poses a small set of Gaussian primitives and maximizes their closed-form overlap with the image SoG, using the hand mask to gate occluded support.
This gives a region-based alignment signal with spatial extent, so the tracker still receives useful gradients under small pose errors, low texture, and partial occlusion.

This is different from three natural alternatives.
First, mesh rendering and dense Gaussian RGB losses depend on view-consistent generated texture.
As shown in~\cref{fig:gaus_vs_sog}, the generated asset often has useful geometry but unreliable front/back or side-specific texture because the multi-view reconstruction is not explicitly constrained by calibrated cameras.
Photometric render-and-compare objectives can therefore optimize toward texture artifacts rather than pose.
Second, sparse optical flow loses tracks under hand occlusion, fast rotations, and large appearance changes.
Note that for the optical-flow baseline, we project object surface points using the current pose, track their 2D locations with iterative Lucas--Kanade\cite{lucas1981iterative} flow, and optimize the rigid object pose so that the projected 3D points follow the tracked 2D positions.
\Cref{tab:ablation} reports the mesh/Gaussian rendering and optical-flow variants, which we outperform with GraG's SoG formulation.

\subsection{Ablations}
\label{supp:ablations}
We conduct ablations on four sequences: ABF14 and GPMF12 from HO3Dv3,  P0011\_0c2d00ed and P0011\_2255f410 from HOT3D.
Unless stated otherwise, we keep the same training schedule and hyperparameters as the full model (including all Stage~3 loss weights and the same number of optimization steps per window).
\Cref{tab:ablation} summarizes the effect of key components: (i) keyframe selection for canonical object reconstruction (Stage~1), (ii) video-level pose tracking with frozen shape and temporal guidance (Stage~2), and (iii) our SoG-based refinement and prior losses (Stage~3).

\paragraph{Attributing gains beyond foundation models.}
The ``only foundation models'' row keeps the SAM3D object reconstruction and Dyn-HaMR hand motion, but removes Stage~3 refinement.
This variant has the same reconstructed object quality, but much worse hand-object alignment ($CD_h{=}20.62$), showing that the final tracking stage is necessary beyond the pretrained priors.
The ``optimization w/o SoG refinement'' row keeps the same Stage~3 optimization and the same geometric/contact priors, but removes the SoG image-alignment term.
Its improvement over ``only foundation models'' shows that the priors correct a large part of the hand-object depth error, while the remaining gap to the full model shows the contribution of SoG tracking.

\paragraph{Full-benchmark SoG ablation.}
\Cref{tab:ablation} ablates SoG on four sequences; here we repeat the priors-only vs.\ priors+SoG comparison on the full benchmark to show the effect of the SoG alignment term.
These averages are over all 18 sequences from HO3Dv3 and 18 sequences from HOT3D, so they are not directly comparable to the four-sequence numbers in \cref{tab:ablation}.
Adding SoG lowers $CD_h$ from $7.66\pm4.93$ to $5.02\pm1.47\,$cm on HO3Dv3 and from $12.11\pm6.39$ to $9.37\pm4.68\,$cm on HOT3D, and tightens the spread across sequences on both.
The variance is larger on HOT3D, as expected from its stronger camera motion and heavier occlusion.
As in the four-sequence study, the priors carry the hand-relative accuracy while SoG enables efficient tracking.

\paragraph{Alternative tracking objectives.}
The mesh, dense Gaussian, and optical-flow rows isolate whether the SoG objective itself is needed.
Mesh and dense Gaussian rendering use the same generated object asset in a direct render-and-compare objective, while optical flow uses Lucas--Kanade tracks of projected object surface points.
All three alternatives increase $CD_h$ relative to full GraG.
They are also slower than GraG, which is consistent with our goal of using the Gaussian asset as a compact tracking template rather than as a dense renderer.

\paragraph{Loss terms.}
Removing contact produces the largest degradation in $CD_h$, confirming that contact is central for resolving hand-object relative depth under occlusion.
Depth, silhouette, and visibility gating provide smaller but consistent gains by anchoring metric scale, aligning object masks, and preventing occluded object Gaussians from matching hand pixels.

\subsection{MagicHOI Result Discrepancy}
We observe a notable variance between the performance of MagicHOI in our experiments compared to the results reported in their original publication. This discrepancy is primarily attributed to differences in input sequence length and domain robustness. While the original MagicHOI study evaluates models on short, curated clips (2–3 seconds), our evaluation utilizes the original HO3D sequence lengths (20–30 seconds).
Also, MagicHOI appears sensitive to temporal scale and viewpoint; it exhibits significantly lower performance when applied to longer, uncurated sequences and egocentric perspectives.

\subsection{Additional Limitation Details}
\label{more_limit}
Our method uses Depth Anything 3 pointmaps as aggregate metric-depth anchors.
Our depth and contact formulation reduces the effect of moderate object-depth errors: the object is not anchored only to its own predicted depth, but is also coupled to the hand depth in frames with contact.
This improves hand-object alignment in cases where the object depth alone would place the object too far from the hand.
However, severe pointmap failures can still affect the reconstruction.
When the pointmap gives inconsistent metric depths for both the hand and object, or when masks/contact cues are unreliable, the available cues may agree on an incorrect metric scale or translation.
In such cases, refinement can still produce an inaccurate hand-object relative pose.

\begin{figure*}[t]
  \centering
\includegraphics[height=0.68\paperheight]{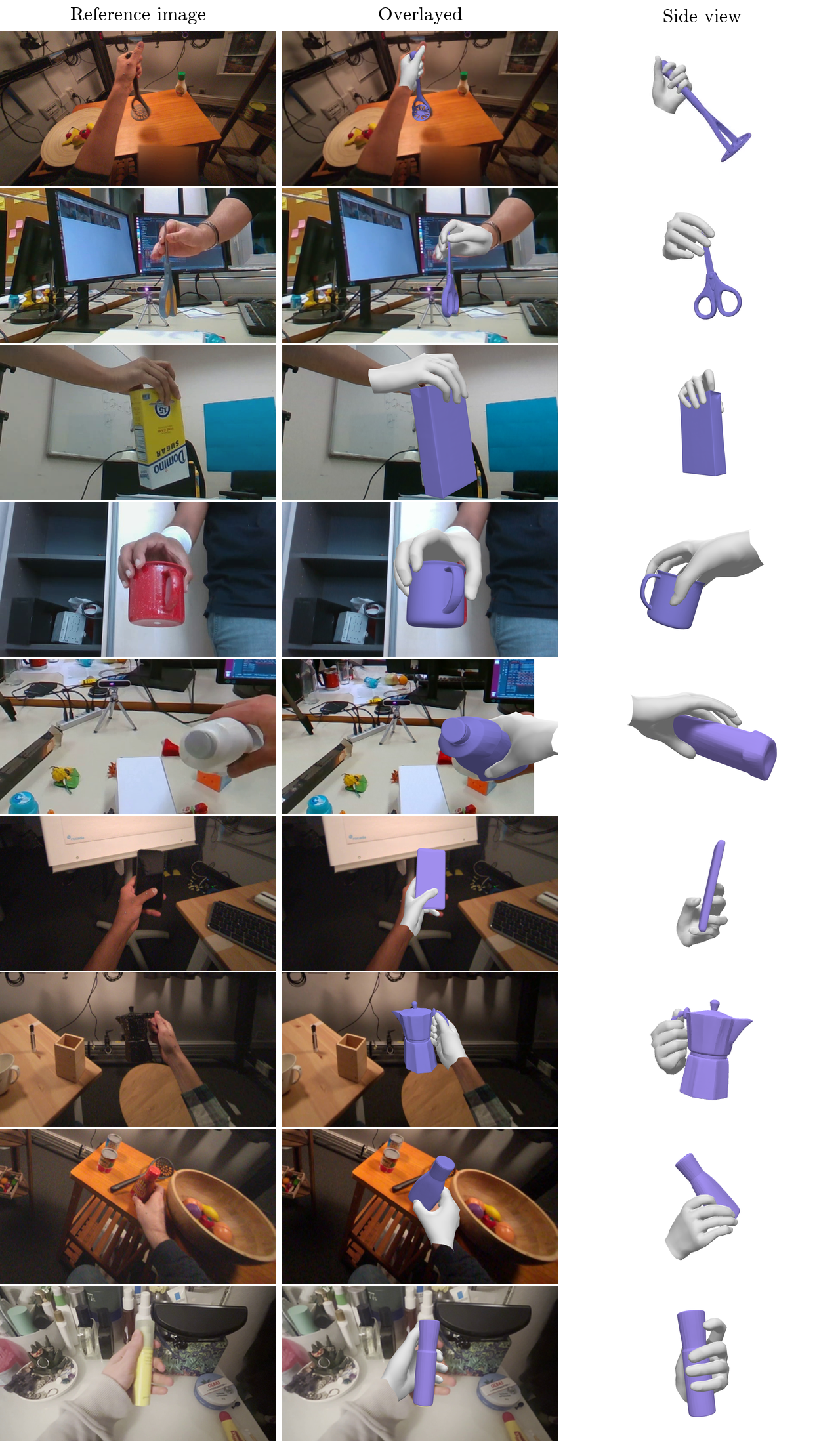}
   \caption{\textbf{More qualitative results.} 
   We overlay rendered hand-object meshes on the input images for qualitative inspection and also show side views of the interactions.
   }
   \label{fig:fig_additional}
\end{figure*}

\end{document}